\documentclass{article}
\usepackage{graphicx} 
\usepackage{amsfonts} 
\usepackage[
sorting=none,
backend=biber,natbib=true,
maxbibnames=99,
style=numeric,
]{biblatex}
\usepackage{pifont}
\newcommand{\xmark}{\ding{55}}%
\usepackage{amsmath}
\usepackage{booktabs}
\usepackage{breqn}
\usepackage{xcolor}
\usepackage{authblk}
\usepackage{lineno}
\usepackage[
    separate-uncertainty=true,
    multi-part-units=single
]{siunitx}
\newcommand{\E}{{\cal E}}
\newcommand{\Ss}{{\cal S}}
\newcommand{\taum}{\tau_{\rm m}}
\newcommand{\tausyn}{\tau_{\rm s}}
\newcommand{\tkspk}{t_{k}^{{\rm spike}}}
\newcommand{\tkev}{t_{k}^{{\rm event}}}
\newcommand{\dt}{{\rm d}t}
\title{Efficient Event-based Delay Learning in Spiking Neural Networks}

\author[1,2]{Balázs Mészáros}
\author[1]{James C. Knight}
\author[1]{Thomas Nowotny}
\affil[1]{Sussex AI, School of Engineering and Informatics, University of Sussex, Brighton, United Kingdom}
\affil[2]{The Alan Turing Institute, London, United Kingdom}
\date{}                     
\begin{document}
 
\maketitle
\begin{abstract}
    Spiking Neural Networks (SNNs) compute using sparse communication and are attracting increased attention as a more energy-efficient alternative to traditional Artificial Neural Networks~(ANNs). While standard ANNs are stateless, spiking neurons are stateful and hence intrinsically recurrent, making them well-suited for spatio-temporal tasks. However, the duration of this intrinsic memory is limited by synaptic and membrane time constants. Delays are a powerful additional mechanism and, in this paper, we propose a novel event-based training method for SNNs with delays, grounded in the EventProp formalism which enables the calculation of exact gradients with respect to weights and delays. Our method supports multiple spikes per neuron and, to the best of our knowledge, is the first delay learning algorithm to be applied to recurrent SNNs. We evaluate our method on a simple sequence detection task, as well as the Yin-Yang, Spiking Heidelberg Digits, Spiking Speech Commands and Braille letter reading datasets, demonstrating that our algorithm can optimise delays from suboptimal initial conditions and enhance classification accuracy compared to architectures without delays. We also find that recurrent delays are particularly beneficial in small networks. Finally, we show that our approach uses less than half the memory of the current state-of-the-art delay-learning method and is up to $26\times$ faster.
\end{abstract}
\section{Introduction}
Artificial Neural Networks (ANNs) have gained immense popularity and seen significant improvements over the past decade. However, commonly used models are very energy intensive \cite{patterson2021carbon}, whereas the human brain has an estimated power budget of only \SI{20}{\watt}~\cite{sokoloff1992brain}. It might, therefore, be beneficial to again turn to neuroscience for inspiration. One reason for the brain's better efficiency is that neurons in the brain transmit sparse binary events called spikes, while the units in ANNs typically communicate real-valued activation values densely in time and space. 
Spiking Neural Networks (SNNs) leverage the sparse communication patterns of biological neurons for machine learning (ML) and are particularly efficient on 
neuromorphic systems designed to provide efficient hardware platforms for brain-like computing~\cite{schuman2017survey,james2017historical,thakur2018large}. 
Like ANNs, SNNs are universal function approximators, suggesting they could enable an energy-efficient future for ML. Therefore, researchers are increasingly focusing on training SNNs on popular ML tasks, for instance, in the fields of computer vision~\cite{sengupta2019going} and natural language processing~\cite{zhuspikegpt}. 
Since individual spiking neurons rely on hidden temporal dynamics, they have `implicit recurrence' so, even SNNs with feedforward architectures, have theoretical advantages for temporal processing over feedforward networks of stateless units. However, despite this potential, achieving or beating the performance of ANNs using SNNs remains a significant challenge.

The most commonly used training algorithm in ANNs is gradient descent. However, the non-differentiable transitions at spike times cause mathematical complications when calculating gradients in SNNs. To overcome this issue, some researchers do not train SNNs directly but instead train ANNs and then transfer the weights to an SNN for inference~\cite{rueckauer2017conversion,stanojevic_high-performance_2024}.
However, because this approach typically uses spike counts to represent the activations of ANN units, it does not fully leverage the potential energy savings of sparse spiking in SNNs. While there are more efficient encoding alternatives \cite{stockl2021optimized}, one time step in the ANN is still mapped to many timesteps in the SNN and the efficiency of SNNs is not leveraged at training time. 

Another popular solution is to discretise the network dynamics and use Backpropagation Through Time (BPTT) with `surrogate gradients' to smooth the threshold function in the backward pass~\cite{bohte_error-backpropagation_2002,neftci2019surrogate}. However, this method requires storing neuron state variables at every time step for the backward pass, meaning that memory requirements scale linearly with sequence length. This limits the maximum number of time steps in a trial to a few hundred. Furthermore, this method also does not exploit the increased efficiency of sparse spiking during the backward pass.

\citet{bohte_error-backpropagation_2002} were the first to show how to calculate exact gradients in SNNs, by providing recursive relations for the gradient that can be implicitly computed using backpropagation. 
 Alternatively, with some constraints on the time constants of neurons, analytic expressions for the time of the next spike of a leaky integrate-and-fire (LIF) can be derived and differentiated~\cite{goltz2021fast,comsa2020temporal}, enabling the calculation of gradients.
However, more generally, neurons in SNNs exhibit hybrid dynamics (a longstanding focus in optimal control theory~\cite{barton2002modeling}), combining continuous changes between spikes with discontinuous state transitions at spike times. The link between neural network training and optimal control has been well established~\cite{selvaratnam2000learning}, and the adjoint method -- a staple of optimal control -- has been used to derive gradients for smoothed spiking neuron models without reset~\cite{huh2018gradient}.
Also using the adjoint method, \citet{wunderlich2021event}  developed the EventProp algorithm for calculating exact gradients in SNNs of integrate-and-fire neurons and `exponential synapses'. The backward pass in EventProp combines a system of ordinary differential equations for the adjoint variables of the neuron dynamics with purely event-based backward transmission of error signals at spike times, making the best use of the hybrid nature of SNNs. \citeauthor{wunderlich2021event} tested their method on latency-encoded MNIST~\cite{lecun1998gradient} and the Yin-Yang datasets~\cite{kriener2022yin}. 
More recently, \citet{nowotny2022loss} extended EventProp to the more challenging benchmarks of the Spiking Heidelberg Digits (SHD) and Spiking Speech Commands~\cite{cramer2020heidelberg}, using loss shaping to overcome issues caused by exact gradients not containing information about spike creation and deletion. 
The time and space complexity of EventProp enables very efficient GPU training of large models on long sequences~\cite{nowotny2022loss}, hardware-in-the-loop training~\citep{pehle_event-based_2023}, and even training on neuromorphic hardware~\cite{bena2024event}. 

Spiking neurons' implicit recurrence is characterised by temporal parameters such as the membrane and synaptic time constants. Research has shown that optimising these~\cite{fang2021incorporating,perez2021neural} can enhance network performance. Delays are another mechanism for temporal processing, and recent work indicates the utility of learnable delays for temporal tasks~\cite{hammouamrilearning, sun2023learnable}, 
In biological neural networks, synaptic delays arise naturally due to the spatial structure of the network and can be modified to facilitate coincidence detection~\cite{seidl2010mechanisms} and learning~\cite{bengtsson2005extensive}. From a computational perspective, the inclusion of delays has been shown to significantly increase network capacity~\cite{izhikevich2006polychronization} and  \citet{maass1999complexity} demonstrated that an SNN with $k$ adjustable delays can compute a much richer class of functions than a network with $k$ adjustable weights. Furthermore, neuromorphic systems such as SpiNNaker~\cite{furber2014spinnaker} and Loihi~\cite{davies2018loihi} are specifically designed to accommodate synaptic delays, so that SNNs with delays can still be efficiently deployed.
Adding delays to SNNs has recently gained popularity, with several models treating them as learnable parameters and obtaining state-of-the-art performance on classification tasks~\cite{hammouamrilearning, sun2023learnable, deckers2024co}. \citet{grappolini2023beyond} even showed that networks can be trained to comparable performance with pure synaptic delay learning. However, most of these methods are based on surrogate gradients~\cite{hammouamrilearning, shrestha2018slayer,sun2023learnable}, which do not allow the event-based nature of SNNs to be exploited at training time, and some use temporal convolutions to implement delays~\cite{hammouamrilearning, wang2019delay}, which results in large overheads in memory and computation. DelGrad~\cite{goltz2024delgrad} was the first exact gradient-based delay learning method, but so far it has only been implemented in feedforward networks where each neuron only emits one spike per trial. No prior work has implemented delay learning for recurrent connections.

Here, we extend EventProp to incorporate heterogeneous and learnable delays and implement our extended version in mlGeNN~\citep{Turner2022,knight2023easy} -- a spike-based ML library built on the GPU-optimized GeNN simulator~\citep{Yavuz2016,Knight2018,knight2021pygenn}.
GeNN generates GPU kernels for efficiently simulating networks of neurons which communicate with sparse events using a \emph{hybrid} simulation strategy where the forward and backward dynamics of each neuron are updated every timestep, but synapses are only updated in timesteps where their presynaptic neurons produce a spike.
This hybrid strategy differs from the purely time-driven approach (typically used to implement SNNs with standard ML libraries), where neurons and synapses are updated every timestep and also from purely event-based simulators where both neurons and synapses are only updated at spike times. All three approaches have advantages and disadvantages.
While a purely timestep-based approach is inherently wasteful, standard ML libraries counteract the inherent inefficiency with highly GPU-optimised matrix multiplication routines to multiply neuron outputs with weights.
Pure event-based simulators make efficient use of the event-based nature of SNNs and produce precise spike times, unconstrained by a timestep grid, but this comes at a cost. 
Only a limited subset of neuron models~\citep{brette_event-driven_nodate,brette_exact_2007} have dynamics that can be directly interpolated between events, algorithms and data structures become increasingly complicated if recurrent connectivity and delays are required~\citep{Brette2007} and, the perceived computational advantage of event-based simulation dwindles rapidly as the frequency of events increases~\citep{Morrison2005}.
For example, each input in the SHD dataset fires at approximately \num{10} spikes per second. If we consider a hidden neuron densely connected to these inputs, it receives \num{7000} spikes per second, meaning that an event-based neuron would have to update $7\times$ more frequently than one simulated using a \SI{1}{\milli\second} timestep.

\citet{nowotny2022loss} described the initial GeNN implementation of EventProp, which has subsequently been implemented as a `compiler'~\citep{knight2023easy} for mlGeNN.
Here, we have added delays to the mlGeNN EventProp compiler, enabling the easy exploration of delay learning in a wide range of network architectures.
When using our method, increasing the range of delays only requires enlarging a \emph{per-neuron} buffer, resulting in a much lower memory overhead than convolution-based approaches and thus allowing efficient handling of long delays.
Our method further allows the outputs of neurons to feed into general spike- and/or voltage-dependent loss functions, offering great flexibility in designing training objectives.
Our approach outperforms prior work using EventProp on the SHD and SSC datasets~\cite{nowotny2022loss} -- achieving superior performance with almost $5\times$ fewer parameters.
Additionally, we demonstrate a speedup of up to $26\times$ and memory savings of over $2\times$ compared to surrogate-gradient-based dilated convolutions implemented in PyTorch~\cite{hammouamrilearning}.

\begin{figure}
    \centering
    \includegraphics[width=0.6\linewidth]{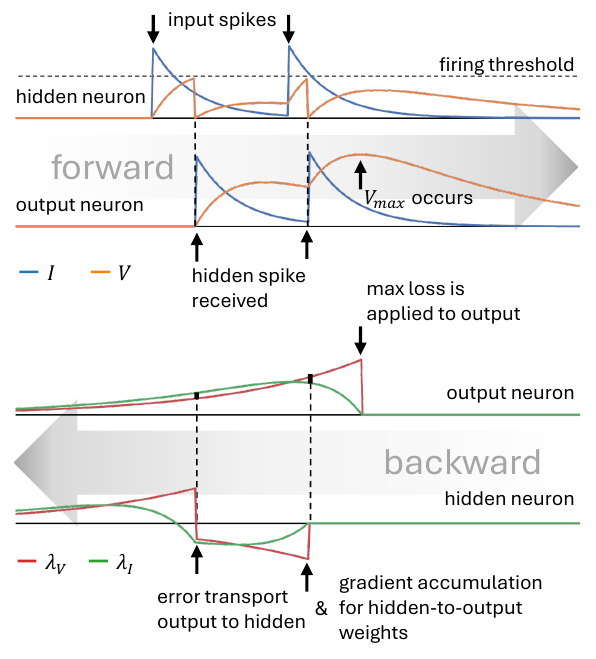}
    \caption{Illustration of the original EventProp formalism without delays. In a minimal example, a network has input neurons, one hidden layer and an output layer. Input spikes cause instantaneous jumps in the hidden $I$ variable (blue lines), which drives the $V$ variable (orange lines). When $V$ reaches the firing threshold, it is reset and spikes are emitted, which instantaneously jump the $I$ variable of the output neurons. This forward pass is followed by a backward pass, where the ``blame'' of each weight for the eventual loss is calculated. The adjoint variables $\lambda_V$ and $\lambda_I$ are proxies of this blame. The calculation occurs backwards in time. Here we illustrate the use of a readout and loss that are based on the maximum voltage of the output neurons. Accordingly, the loss causes a jump in the output $\lambda_V$ variable at the time when the maximum output voltage occurred in the forward pass. This blame is then transported as jumps of $\lambda_V$ of hidden neurons at the times when hidden spikes had occurred. Gradient updates to the hidden-to-output weights occur at the same time. Note that the plots are for illustrative purposes only and not to scale, other than matching pairs of $\lambda_V$ and $\lambda_I$ being at the same scale.}
    \label{fig:basic_explain}
\end{figure}

\begin{figure}
    \centering
    \includegraphics[width=0.6\linewidth]{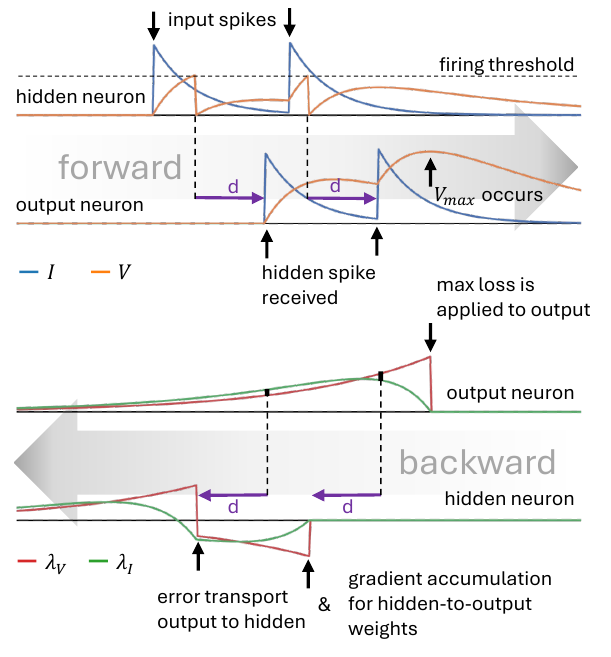}
    \caption{Illustration of the extended EventProp algorithm for SNNs with delays. In essence, the forward pass works in the same way as for networks without delays (Fig. \ref{fig:basic_explain}), except that the jump in post-synaptic $I$ variables occurs with a delay $d$. In the backward pass, this translates into backwards transport of blame at the original spike times but transporting values of post-synaptic adjoint variables at a delayed time (in forward time). Gradients with respect to weights and delays are accumulated at the same saved spike times and based on the same delayed quantities.}
    \label{fig:explain_delay}
\end{figure}

\begin{table} 
\label{tab:dynamics}
\begin{tabular}{lll} 
\toprule 
{\bf Free dynamics} & {\bf Transition } & {\bf Jumps at transition} \\ & {\bf condition} & \\ 
\midrule 
Forward: \\ $\taum \dot{V} = -V + I$ & $(V^-)_{n(k)} \Big|_{\tkspk} - \vartheta = 0$, & $(V^+)_{n(k)}\Big|_{\tkspk} = 0$ \\  $\tausyn \dot{I}= -I$ & $\big(\dot{V}\big)_{n(k)} \Big|_{\tkspk} \neq 0$ & $(I^+)_m \Big|_{\tkspk \textcolor{red!70}{+d_{mn}(k)}} =(I^-)_m \Big|_{\tkspk \textcolor{red!70}{+d_{mn(k)}}}+w_{mn(k)}e_{n(k)}$ \\ 
Backward: \\ $\taum \lambda_V' =-\lambda_V-\frac{\partial l_V}{\partial V}$, & $t-\tkspk =0$, & $(\lambda_V^-)_{n(k)} = \left. \left[\frac{(\dot{V}^+)_{n(k)}}{(\dot{V}^-)_{n(k)}}(\lambda_V^+)_{n(k)}+\frac{1}{\taum(\dot{V}^-)_{n(k)}}\left[\frac{\partial l_p}{\partial \tkspk }+ l_V^--l_V^+\right]\right]\right|_{\tkspk }$
\\  
$\tausyn \lambda_I' =-\lambda_I+\lambda_V$ & for any $k$  & $+\left. \left[\frac{1}{\taum(\dot{V}^-)_{n(k)}}\right]\right|_{\tkspk } \sum_{m}w_{mn(k)} \left.\left[(\lambda_V^+ -\lambda_I^+)_m\right]\right|_{\tkspk \textcolor{red}{+d_{mn(k)}}}$ \\ 
\toprule 
{\bf Gradient updates} \\ 
\midrule 
\multicolumn{3}{p{16cm}}{\parbox{8cm}{Weight learning: \\
 $\displaystyle \frac{{\rm d}\mathcal{L}}{{\rm d}w_{ji}} =-\tausyn \sum_{\parbox{0.9cm}{\scriptsize $\tkspk$ \parbox{0.9cm}{from\ $i$}}}(\lambda_I)_j\Big|_{\tkspk \textcolor{red!70}{+d_{ji}}}$}
\parbox{8cm}{Delay learning: \\
$\displaystyle \textcolor{red!70}{\frac{{\rm d}\mathcal{L}}{{\rm d}d_{ji}} =- w_{ji}\sum_{\parbox{0.9cm}{\scriptsize $\tkspk$ \parbox{0.9cm}{from\ $i$}}} (\lambda_I-\lambda_V)_j\Big|_{\tkspk+d_{ji}}}$ }}
\label{tab:system}
\end{tabular} 
\caption{
Forward and backward propagation of a Leaky Integrate-and-Fire~(LIF) neuron. $V$ and $I$ are the membrane potential and input current and $\lambda_V$ and $\lambda_I$ the corresponding adjoint variables. $\taum$ and $\tausyn$ are the membrane and synaptic time constants. $w_{mn}$ is the synaptic weight and $d_{mn}$ is the synaptic delay from neuron $n$ to neuron $m$. $\vartheta$ is the firing threshold. The dot denotes the derivative with respect to time, and the prime the derivative backwards in time. Superscript ``-'' and ``+'' denote the values before and after a transition. $\tkspk$ denotes the $k^{\rm th}$ spike and $n(k)$ the index of the neuron that fired this spike. $l_p$ and $l_V$ define the shape of the loss function. Without the transition condition (in other words, with $\vartheta=\infty$), we arrive at the Leaky Integrator~(LI) neuron. We highlight all our additions in red.}
\end{table}

\section{Results}
The EventProp algorithm is an application of the adjoint method for calculating sensitivities in hybrid dynamical systems \citep{galan1999parametric,Rozenvasser1967} to the problem of calculating the gradient of a loss function in an SNN. From an ML perspective, it can be considered an event-driven form of the popular BPTT gradient descent-based learning in RNNs. The difference is that the gradient computation employs a hybrid approach: the derivatives are determined through both continuous differential equations and discrete state transitions of adjoint variables that occur at saved spike times.

Intuitively, in an SNN, synaptic weights can only affect the network activity, and hence cause loss, at times when a spike is transmitted. The adjoint variables track the potential loss caused by each neuron, and their value at the time of a spike quantifies how much loss that spike contributed to the overall loss, essentially assigning responsibility or ``blame'' to both the spike event and the synaptic weights involved. For output neurons, blame for loss is directly added to the adjoint variable corresponding to the membrane dynamics (jump in $\lambda_V$ of the output neuron where $V_{\rm max}$ occurred in Figure \ref{fig:basic_explain}), while for internal neurons, it propagates backward from downstream connections (jumps in $\lambda_V$ of the hidden neuron at saved spike times in Figure \ref{fig:basic_explain}). The gradient for each synaptic weight ($w_{ji}$) accumulates across pre-synaptic firing events, capturing the moments when that connection influenced the post-synaptic neuron's behaviour and consequently affected network performance. The differential equations governing the adjoint variables precisely track how input fluctuations drive membrane changes that ultimately impact the loss -- whether by triggering spikes or directly affecting the output neuron's contribution to the loss function.

Here, we derive an extended EventProp formalism that extends the original algorithm in both, that it can be applied to SNNs with delays and that it enables learning of suitable delays, i.e. to calculate gradients of a loss function with respect to delays.
Although the EventProp formalism accommodates various neuron models~\cite{pehle2021adjoint}, for simplicity, we will describe it for the LIF neurons and exponential current-based synapses employed by \citet{wunderlich2021event}. 

The forward pass of the SNN is described by first-order ordinary differential equations for the dynamics of the current $I$ and voltage $V$; and discontinuous jumps in the variables at the occurrence of spikes (see table~\ref{tab:system}). Note that this only differs from the forward pass described by \citet{wunderlich2021event} due to the delay $d_{mn}$ between neuron $n$ and neuron $m$ which causes the $k$-th jump in the network in the current $I$ of neuron $m$ caused by a spike in neuron $n$ at time $\tkspk$ to occur at time $\tkspk + d_{mn}$, see also Figure \ref{fig:explain_delay}, top.
If all $d_{mn}$ are zero, this reverts back to the original EventProp formalism.

To obtain the backward pass for our network with delays, we take the derivative of the loss function with respect to a weight, $w_{ji}$. The loss function can depend directly on spike timing (which is compatible with the gradient calculation because LIF neuron spike times vary smoothly with weight changes), or it can be expressed as an integral over the voltage $V$ (where the integral effectively smooths out the discontinuities that would otherwise make the gradient calculation difficult when spikes occur). 
\begin{equation} 
    \frac{{\rm d}\mathcal{L}}{{\rm d}w_{ji}} =\frac{{\rm d}}{{\rm d}w_{ji}}\left[l_p(\Ss) + \int_{t=0}^{T} l_V(V,t) \dt \right] , 
\end{equation}
where $l_p$ is the loss term that depends on the set of output spike times $\Ss \equiv \{\tkspk \; | \; k=1, \ldots, N_{\rm spike}\}$, and $l_V$ is the voltage-dependent loss. Following the adjoint method, we then add Lagrange multipliers $\lambda_I$ and $\lambda_V$, multiplied by the continuous dynamics, which can be interpreted as dynamic constraints,
\begin{equation} 
    \frac{{\rm d}\mathcal{L}}{{\rm d}w_{ji}} =\frac{{\rm d}}{{\rm d}w_{ji}}\left[l_p(\Ss) + \int_{t=0}^{T} \left[l_V(V,t) +\lambda_V f_V +\lambda_I f_I \right] \dt \right] , \label{eqn:loss_dev2}
\end{equation}
where $f_V \equiv \taum \dot{V} + V - I$ encodes the voltage dynamics constraint and ${f_I \equiv \tausyn \dot{I} + I}$ the current dynamics constraint.

The essence of the original EventProp derivation was to split the integral in (\ref{eqn:loss_dev2}) at the $N_{\text{spike}}$ spike times $\tkspk$ when the jumps occur. Between the jumps, everything is well-defined and the standard adjoint method is easily applied -- resulting in the backward dynamics for the adjoint variables. 
With some work (see derivations by \citet{wunderlich2021event}), the values of adjoint variables before and after jump times in the backward pass can then be defined so that the remaining expression for the gradient becomes a simple sum over $\lambda_I$ values at spike times, leading to an event-based weight update rule: 
\begin{align}
    \frac{{\rm d}\mathcal{L}}{{\rm d}w_{ji}} &=-\tausyn \sum_{\parbox{0.9cm}{\scriptsize $\tkspk$ \parbox{0.9cm}{from\ $i$}}} (\lambda_I)_j \Big|_{\tkspk}.
\end{align}

We apply a similar approach here, but in our network with delays, spike emission and arrival times become separate events.
We address this by extending the set of spike times to the set of all event times that include both spike emission times $\tkspk$ and spike arrival times $\tkspk + d_{mn(k)}$, where $n(k)$ is the index of the neuron that fired the $k^{\text{th}}$ spike: 
\begin{align}
    \E \equiv \Ss \cup \{\tkspk+d_{m,n(k)} \; | \; k= 1,\ldots,N_{\text{spike}}, m= 1,\ldots,N\}.
\end{align}
We denote the elements of this set as $\tkev \in \E$ and assume that they are ordered such that $\tkev  \leq t_{k'}^{\text{event}}$ for $k < k'$.
We can then proceed in the same way as in the original EventProp derivation.
The resulting backward pass becomes computable because all delays are nonnegative, meaning that by the time we compute the adjoint variables for the spiking neuron $i$ at time $t$, the adjoint variables for the postsynaptic neurons $j$ receiving the spike at time $t+d_{ji}$ will have already been calculated before, in backward time (see Figure \ref{fig:explain_delay}, bottom).

\begin{figure}
    \centering
    \includegraphics[width=0.7\linewidth]{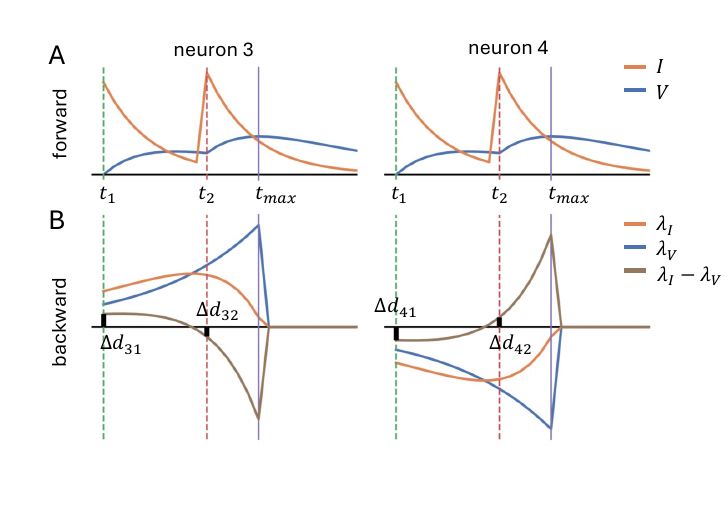}
    \caption{Illustration of the learning updates in the sequence detection task. A) Voltage $V$ and current $I$ in the forward pass. B) Adjoint variables $\lambda_V$ and $\lambda_I$. The loss is injected into $\lambda_V$ at $t_{\text{max}}$ when the output voltage reached maximum in the forward pass. This then propagates into $\lambda_I$ and eventually leads to delay updates at the saved spike times $t_1,t_2$. For neuron $3$, the update to $d_{32}$ at $t_2$ is negative and the update to $d_{3,1}$ at $t_1$ positive, meaning that (subject to delays being non-negative) the excitatory postsynaptic potential (EPSP) from neuron 2 is moved to earlier and the EPSP from neuron 1 to later, moving them close together. For neuron 4, the opposite is the case, where the updates are such that the EPSPs are moved apart. This is exactly what is needed to increase the maximal output voltage of neuron 3 and decrease the maximal output of neuron 4. When the other input class is active, all roles are inverted, again leading to the correct delay updates.}
    \label{fig:sequence_explain}
\end{figure}
\begin{figure}
    \centering
    \includegraphics[width=\linewidth]{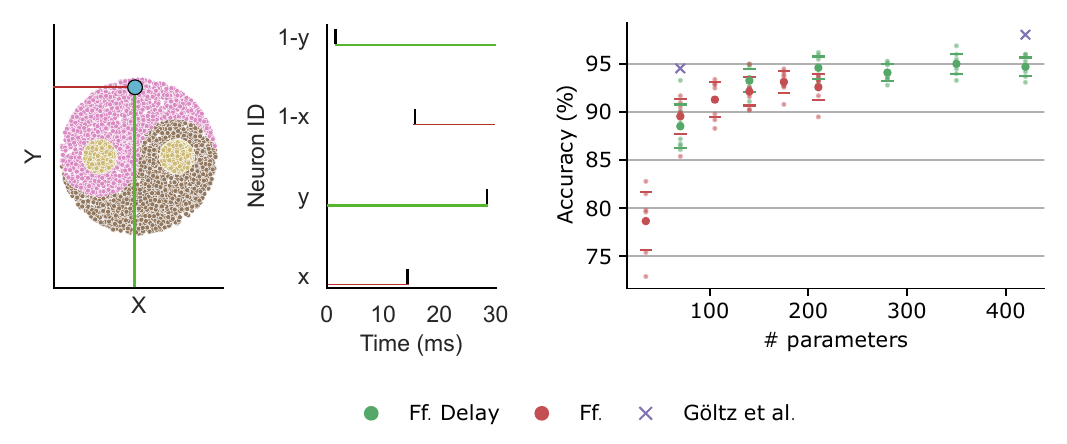}
    \caption{\textbf{Left} The Yin-Yang (YY) dataset, with temporal encoding of example datapoint highlighted by a blue dot. \textbf{Right}
     We generate separate training, validation, and test sets with 5000, 1000 and 1000 examples respectively; and report the test performance using the model which performed best on the validation set. We look at feedforward networks with and without delays. The purple crosses shows the results reported by \citet{goltz2024delgrad}.  The points show average accuracy, and the error bars show standard deviation over 8 runs. We also show all individual results as smaller data points.}
    \label{fig:yy}
\end{figure}

After extensive calculations (section \ref{methods}), we arrive at a formula that remains fully event-based for synaptic actions in the backward pass and thus can be efficiently computed,
\begin{equation}
    \frac{{\rm d}\mathcal{L}}{{\rm d}w_{ji}} =-\tausyn \sum_{\parbox{0.9cm}{\scriptsize $\tkspk$ \parbox{0.9cm}{from\ $i$}}}(\lambda_I)_j\Big|_{\tkspk +d_{ji}} .
\end{equation}

Using the same approach, we can also derive the derivative of the loss function with respect to the delays $d_{ji}$. Remarkably, the derived backward dynamics of the adjoint variables remain the same, meaning that using the exact same  $\lambda_I$ and $\lambda_V$ dynamics in the backward pass, we {\em can also perform gradient descent on synaptic delays in an event-based manner}. The resulting formula for the delay gradients is
\begin{equation}
    \frac{{\rm d}\mathcal{L}}{{\rm d}d_{ji}} =- w_{ji}\sum_{\parbox{0.9cm}{\scriptsize $\tkspk$ \parbox{0.9cm}{from\ $i$}}} (\lambda_I-\lambda_V)_j\Big|_{\tkspk+d_{ji}}.
\end{equation}
We thus end up with an event-based weight and delay learning algorithm which, we believe, is the first to enable delay learning in networks with multiple recurrently-connected layers and with multiple spikes per neuron.

\subsection{Sequence detection task}

To test the effectiveness of our method, we evaluated it in various machine learning settings. First, we generated a simple binary classification task that can be solved with perfect accuracy using optimal delays. Specifically, we used two LIF neurons connected to two LI neurons (see Table \ref{tab:dynamics}), with all connection strengths set to 1. The task involves two classes:
\begin{itemize}
    \item Class 1: The first input neuron emits a spike at \SI{0}{\ms}, and the second emits a spike at \SI{10}{\ms}.
    \item Class 2: The first input neuron emits a spike at \SI{10}{\ms}, and the second emits a spike at \SI{0}{\ms}.
\end{itemize}
The output of the network is determined based on the maximum voltage reached by the two output neurons, each corresponding to one of the output classes. The class associated with the neuron that reaches the higher voltage is selected as the network's prediction. We can observe that having 10 in the diagonals and 0 everywhere else in our $2\times2$ delay matrix solves the task. We started with the least optimal delay distribution -- delays on the diagonals being 0, and 10 everywhere else --  and, with the learning rate set to 1, we achieve \SI{100}{\percent} accuracy after encountering both examples 6 times. This result demonstrates that, by introducing our delay updates into the learning framework, SNNs become capable of not only coincidence but sequence detection. Figure \ref{fig:sequence_explain} illustrates the gradient calculation in this task. 

\subsection{Yin-Yang dataset}
We also experimented with the Yin-Yang dataset~\cite{kriener2022yin}, which has been tested using both EventProp (without delays)~\cite{wunderlich2021event} and DelGrad~\cite{goltz2024delgrad}, see Figure \ref{fig:yy}. Similarly to DelGrad, we looked at feedforward networks and varied the size of the hidden layer from \num{5} to \num{30}. We initialised all delays at 0, allowing them to evolve during training and trained using the time-invariant mean squared error loss~\cite{goltz2024delgrad}, see \ref{methods} for derivation. Our findings (Figure \ref{fig:yy}) are very similar to DelGrad -- with a fixed number of parameters, networks perform similarly (i.e. halving the number of hidden neurons does not decrease performance if delays are introduced). If the number of parameters is not a constraining factor, training delays \emph{and} weights is always advantageous. 

\subsection{Spiking Heidelberg Digits}
\begin{figure}
    \centering
    \includegraphics[width=\linewidth]{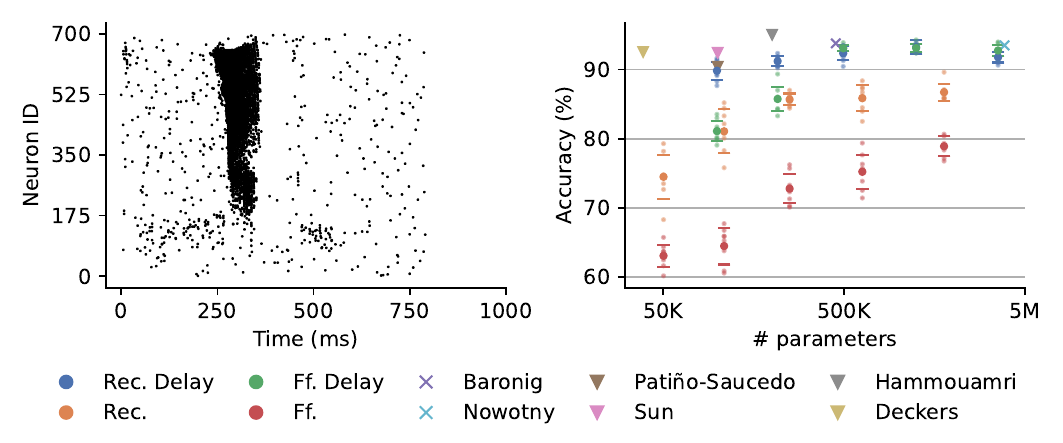}
    \caption{\textbf{Left} An example of a speaker saying ``five'' from the Spiking Heidelberg Digits (SHD) dataset. \textbf{Right} The SHD dataset does not have a validation set; we perform early stopping when training accuracy does not improve for 15 epochs and report the corresponding test accuracy. Ff denotes 
 a model with 2 feedforward hidden layers, and Rec denotes a model with one recurrently connected hidden layer. We implemented models with 128, 256, 512 and 1024 hidden neurons. We also show SOTA results by \citet{baronig2024advancing}, and previous EventProp results by \citet{nowotny2022loss}. The triangles show results of other delay learning methods that appear to have used the test set for validation \citep{hammouamrilearning,sun2023learnable,deckers2024co}. The points show average accuracy, and the error bars show standard deviation over 8 runs. We also show all individual results as smaller data points.}
    \label{fig:shd}
\end{figure}

\citet{nowotny2022loss} achieved state-of-the-art results on SHD with a `delay line' approach, which involved creating 10 copies of the input and cumulatively delaying each by \SI{30}{\ms}. While this architecture achieved high accuracy, it required a large number of parameters, so we instead experimented with learnt delays in the input-to-hidden and hidden-to-hidden connections. For controlling the training dynamics in the hidden population, a target firing rate needs to be set, with the corresponding spike regularisation strength. We kept our target firing rate fixed at 14 spikes per example and treated the regularisation strength as a tunable hyperparameter, which we optimised using 10-fold cross-validation, leaving one speaker out of the training set in each fold. Tuning this parameter was crucial (particularly for networks with recurrent connectivity) and once we identified the best-performing model in cross-validation, we retrained it with the same parameters on the full training set. We enforced early stopping if training accuracy did not improve for 15 epochs. Using this methodology, our best model achieved a training accuracy of \SI{98.47 \pm 0.4}{\percent} and a test accuracy of \SI{93.24 \pm 1.0}{\percent} as depicted in Figure~\ref{fig:shd}. This configuration included 512 hidden neurons with recurrent connections. The feedforward delays were initialised from a uniform distribution in the range of \SIrange{0}{150}{\ms}, while the recurrent delays were all initialised to \SI{0}{\ms}.
The difference between these results and those reported using the `delay line' approach (\SI{93.5 \pm 0.7}{\percent}~\citep{nowotny2022loss}) are not statistically significant ($p=0.442$, t-test, $n=8$), and we achieved them with around 5 times fewer parameters.
We also experimented with different hidden layer sizes and feedforward models. We found that decreasing the hidden neuron number to $256$ does not significantly decrease the accuracy for either architecture. However, if we decrease the number of neurons even further to $128$, we observe a significant drop for the feedforward architecture but not for the recurrent one. Increasing the hidden layer to 1024 neurons leads to overfitting. 

While state-of-the-art models -- such as the work by \citet{hammouamrilearning, deckers2024co,sun2025towards} -- reported higher accuracies, these were obtained using the test set for validation and early stopping, rather than using a separate validation set. \citet{schone2024scalable} mention that they also evaluate in this way to achieve a ``fair comparison to others''.
However, as \citet{baronig2024advancing} argued, not only is this not methodologically ``clean'' but it may also not be entirely fair due to potential overfitting (we note that the highest test accuracy we observed was \SI{95.32}{\percent}). 
Furthermore, we also note that model performance on the SHD dataset is nearing saturation, with the best-performing models achieving an accuracy of around \SI{93}{\percent}. Following \citet{isaksson2008cross} and \citet{nowotny2014two}, we calculated the Bayesian confidence intervals with naive assumptions on error rates. \SI{93}{\percent} accuracy has overlapping confidence intervals with higher accuracies (e.g., \SI{94}{\percent} and \SI{95}{\percent}), indicating that further improvements in accuracy are likely not statistically meaningful given the test set size ($2264$)~\cite{nowotny2014two}.

\subsection{Spiking Speech Commands}
\begin{figure}
    \centering
    \includegraphics[width=\linewidth]{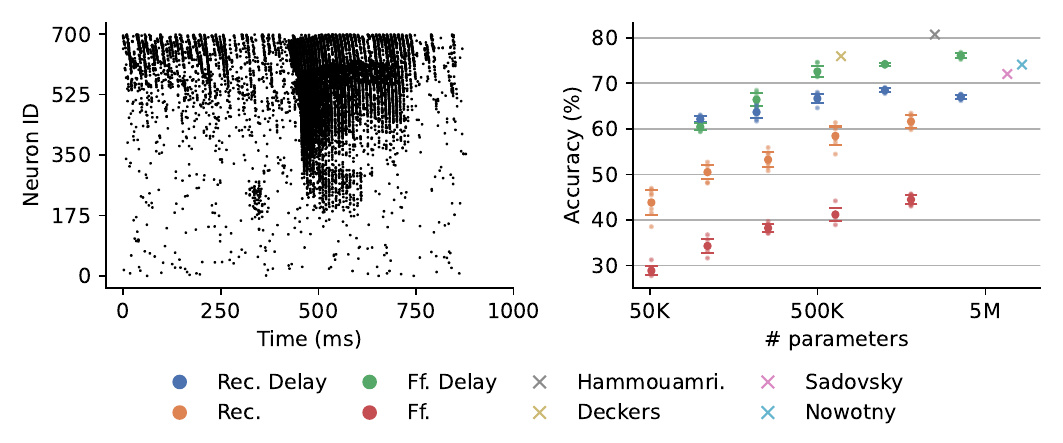}
    \caption{\textbf{Left} An example of a speaker saying ``two'' from the Spiking Speech Commands (SSC) dataset. \textbf{Right} For SSC, which has separate training, validation, and test sets, we apply early stopping when validation accuracy no longer improves and report the corresponding test accuracy. Ff denotes 
 a model with 2 feedforward hidden layers, and Rec denotes a model with one recurrently connected hidden layer. We implemented models with 128, 256 and 512 hidden neurons. Crosses show results from other delay learning models \citep{hammouamrilearning, deckers2024co, sadovsky2023speech}, and we also show previous Eventprop results by~\citet{nowotny2022loss}. The points show average accuracy, and the error bars show standard deviation over 8 runs. We also show all individual results as smaller data points.}
    \label{fig:ssc}
\end{figure}
SSC is significantly more challenging than SHD as the audio recordings were created in noisy environments, and the dataset has more classes. We initially experimented with single recurrent hidden layer architectures similar to those employed by \citet{nowotny2022loss} and, after replacing the delay line inputs with learnable delays, we achieved similar performance. Interestingly, we observed little to no benefit of adding delays to larger networks but, as we decreased the number of hidden neurons, the delays became highly beneficial. These architectures were extremely robust to decreasing the number of hidden neurons.
While many state-of-the-art models use deeper architectures with more hidden layers~\cite{hammouamrilearning, bittar2022surrogate,baronig2024advancing}, we found that deeper architectures with recurrent connections became highly unstable even without delays in the connections.
Therefore, to improve upon previous results, we explored deeper \emph{feedforward} architectures with delays and found the best performing architecture to be a model with 2 feedforward hidden layers. 

Our results are shown in Figure \ref{fig:ssc}. Our best model achieved a training accuracy of \SI{79.6 \pm 1.0}, a validation accuracy of \SI{78.1 \pm 1.0}{\percent} (n=8) and a test accuracy of \SI{76.1 \pm 1.0}{\percent}. We also experimented with smaller models, which as expected, achieved a lower training accuracies. Compared to other SNNs with delays, we observe that we outperform \citet{sadovsky2023speech}, with their test accuracy results at  \SI{72.03}{\percent}. \citet{deckers2024co} introduced a constrained adaptive LIF neuron model to a delayed network and reached \SI{80.23}{\percent} test accuracy. They also tested LIF models, achieving \SI{75.94}{\percent}, which we slightly outperformed.

\subsection{Braille letter reading}
\begin{figure}
    \centering
    \includegraphics[width=\linewidth]{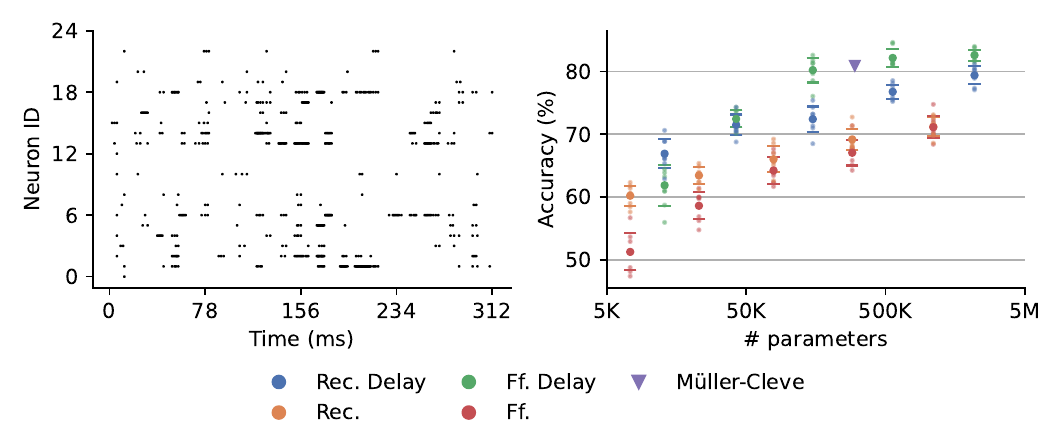}\caption{\textbf{Left} Example of the letter 'S' from the braille letter reading dataset. We split the dataset into training, validation and test set in a ratio of $70:10:20$, respectively. We tune our hyperparameters based on validation results and report the corresponding test accuracy. \textbf{Right} Ff denotes a model with 2 feedforward hidden layers and Rec denotes a model with one recurrently connected hidden layer. We implemented models with 128, 256, 512 and 1024 neurons. The triangle depicts validation results reported by \citet{muller2022braille}. The points show average accuracy, and the error bars show standard deviation over 8 runs. We also show results from all individual results with smaller data points. }
    \label{fig:braille}
\end{figure}
As highlighted by~\citet{walters2025neuromorse}, most neuromorphic benchmarks contain more spatial than temporal information. Although the SHD and SSC datasets \emph{do} contain temporal information, it may not be important enough to require the fine timesteps our approach enables. Therefore, we also evaluated our approach on a braille letter reading dataset~\citep{muller2022braille}. 

We again trained, validated and tested 2-layer feedforward and single-layer recurrent networks with and without delays and with hidden layers of various sizes ($64, 128, 256, 512, 1024$) on the $70:10:20$ training, validation, and test splits provided by \citet{muller2022braille}. The results shown in Figure~\ref{fig:braille} show a similar pattern to the SSC results -- introducing delays in large recurrent networks shows no benefit, but adding them to smaller networks significantly improves performance.

Our best-performing model had two feedforward hidden layers, with 1024 neurons each, and achieved \SI{83.1 \pm 1.5}{\percent} on the test set. This model outperforms the recurrent network with a single hidden layer of 450 neurons and 8 input copies described in the original publication~\citep{muller2022braille}, which obtained a test accuracy of \SI{80.9 \pm 0.3}{\percent}. Our smaller feedforward network with hidden layers of size 256 achieved a test accuracy of \SI{81.0 \pm 0.7}{\percent} so also outperformed \citet{muller2022braille}, with half the number of parameters.
Additionally, \citet{muller2022braille} evaluated their models using an $80:20$ training-test split without a validation set, which is problematic for the same reasons identified in SHD dataset studies. 
\citet{pedersen2024neuromorphic} previously trained on this dataset but simplified it to 7 classes to accommodate the Xylo chip~\citep{bos2023sub}.
We are not aware of any other research involving delay learning on this dataset.

\subsection{Computational performance}
\begin{figure}
    \centering
    \includegraphics[width=\linewidth]{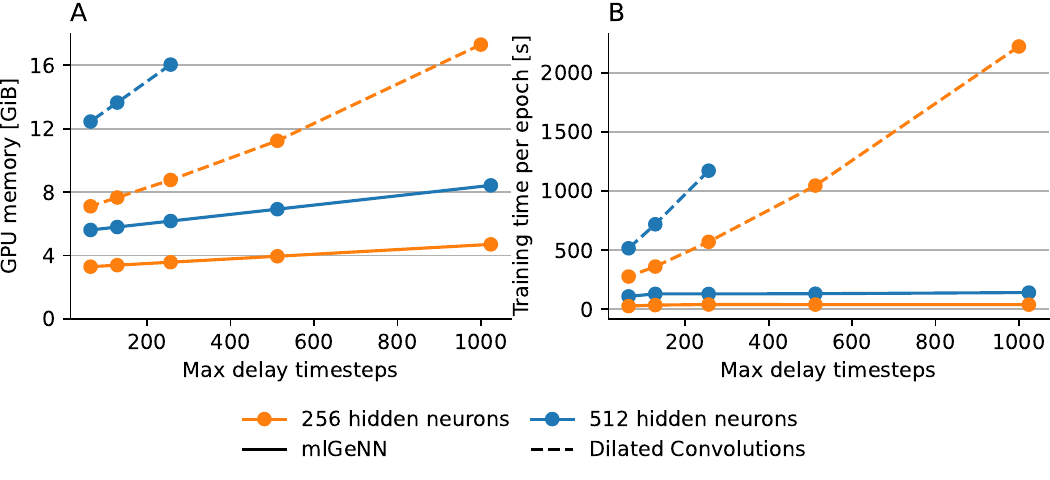}
    \caption{Comparing cost of training networks with delays using EventProp and mlGeNN against Dilated Convolutions~(DC) implemented using SpikingJelly and PyTorch. \textbf{(A)} Peak GPU memory usage. Missing data points indicate where PyTorch ran out of GPU memory. \textbf{(B)} Time to train one SHD epoch. All experiments were performed on a workstation with an NVIDIA RTX A5000 GPU. All models have two feedforward hidden layers and use batch size 256 and \SI{1}{\ms} timesteps.}
    \label{fig:dc_mlgenn}
\end{figure}
Finally, we benchmarked our training procedure against the Dilated Convolution implementation provided by \citet{hammouamrilearning} using PyTorch 2.5.1 and SpikingJelly 0.0.0.0.15~\citep{spikingjelly}.
Because the Dilated Convolution method does not support recurrent delays, we benchmarked feedforward models with 2 hidden layers.
We measured the peak memory utilisation of mlGeNN using the  ``nvidia-smi'' command line tool -- as GeNN allocates memory from CUDA directly -- and of PyTorch using torch.cuda.max\_memory\_allocated -- so details of the memory allocator are disregarded.

The benchmarking results are illustrated in Figure~\ref{fig:dc_mlgenn}. Because increasing the maximum number of delay slots in mlGeNN only involves increasing the size of a \emph{per-neuron} buffer rather than increasing the size of a \emph{per-synapse} kernel, longer delays have much lower memory requirements in mlGeNN (Figure~\ref{fig:dc_mlgenn}A).
Similarly, the increasing computational cost of convolving larger and larger kernels means that training time increases rapidly when using Dilated Convolutions (Figure~\ref{fig:dc_mlgenn}B), whereas with mlGeNN, there is only a very small initial increase in training time as the maximum number of delay timesteps increases. 
Other gradient-based delay learning methods~\cite{wang2019delay,shrestha2018slayer,sun2023learnable} do not use temporal convolutions, so their memory and computational requirements would not grow \emph{as} quickly, but they are all still based on dense BPTT so -- based on benchmarking performed by \citet{nowotny2022loss} -- we would still expect mlGeNN to be significantly faster and use less memory.

The slight increase in mlGeNN training time observed as the maximum number of delay timesteps increases is likely due to the effects of caching on the delay buffers. These buffers are allocated in GPU global memory, so the efficiency of updating them depends on whether they persist in the GPU's caches for long enough for locality in the delays to be exploited.
At the large batch sizes used for these experiments, the complete buffers will not fit in the \SI{6}{\mebi\byte} L2 cache of the RTX A5000 GPU so the effects seen here are likely due to an increase in the rate at which delay buffer data gets evicted from the cache as the size of the buffers increases.

\section{Discussion}
Delays in Spiking Neural Networks have been extensively studied from both machine learning and computational neuroscience perspectives, with recent interest spurred by their ability to improve performance in machine learning applications~\cite{shrestha2018slayer,hammouamrilearning}. Delays can be efficiently implemented on several neuromorphic chips~\cite{davies2018loihi,furber2014spinnaker}, but, as is the case with much of current neuromorphic research, there is a lack of focus on delay \emph{learning} algorithms that could be implemented on future neuromorphic hardware. Instead, many methods rely on arithmetically intensive approaches that are only practical on GPUs, such as convolutions in networks trained with BPTT with surrogate gradients.
Our method combines the best of both worlds: its theoretical foundations enable efficient implementation on neuromorphic hardware (EventProp has already been implemented in SpiNNaker 2~\cite{furber2014spinnaker}), while our GeNN-based implementation takes advantage of readily available GPU hardware. This versatility allows us to address a broader range of platforms while improving performance on complex temporal tasks.


The beneficial scaling of EventProp allows long sequence lengths and hence finer timesteps, which might offer enhanced precision in spatio-temporal tasks. However, leading models on SHD and SSC employ large timesteps of \num{10} or even \SI{25}{\milli\second}~\cite{hammouamrilearning,bittar2022surrogate} and, while these coarser time grids may have been primarily chosen to fit within GPU memory, they could also simplify the tasks by reducing their effective sequence length (assuming high temporal precision is not required). This reflects a broader challenge in SNN research: while neuromorphic architectures are appealing for their energy efficiency and temporal precision, it remains unclear how much temporal precision is needed for any given task and how it is best exploited in practice. 
Timestep length also has additional relevance for delay learning, as delays are discretised to integer multiples of the timestep. However, our previous work indicates that the speech recognition tasks considered here may not require precise delays~\cite{meszaros18learning}. 

Another open question regards initialisation.
While synaptic weights can be initialised in a straightforward manner by sampling from a normal distribution centred around zero, the initialisation of delays is less intuitive. Experimental observations of delay distributions are challenging to interpret~\cite{swadlow1985physiological}, and the optimal distribution may depend heavily on the task. We followed the common approach of initialising delays with values sampled from uniform distributions~\cite{hammouamrilearning}. However, interestingly, we observed that while a few delays grew considerably, most remained relatively short. This distribution might reflect a small-world network structure, where most neurons connect to nearby neighbours (short delays) with only a few long-range connections (long delays). 

As we show across multiple datasets, recurrent delays can offer substantial advantages when implementing networks with stricter size constraints. However, initialising recurrent delay distributions poses additional computational challenges. Initialising feedforward delays within the range $[0,d_{\rm max}]$ is logical, as adding a homogeneous delay $x$ would yield the same outcomes within the range $[x,d_{\rm max}+x]$. However, this symmetry does not extend to recurrent delays, making their initialisation significantly more complex. Recent studies have focused on optimising delay distributions at the network level~\cite{moro2024role}, but the question of layer-specific initialisation remains open.

This work showed the benefit of delays on datasets commonly used to benchmark SNNs. Exploring them in other tasks where they could be particularly beneficial (e.g. sound localisation~\cite{jeffress1948place}, or motion detection~\cite{grimaldi2022learning, grimaldi2023learning}) is an interesting avenue for future work.
\section{Methods} \label{methods}
\subsection{Theory}
\subsubsection{Learning Weights in networks with delay}
We start by defining our two differential equations in the implicit form for the membrane potentials and input currents, respectively.
\begin{align}
    f_V &\equiv \taum \dot{V}+V-I=0 \\
    f_I &\equiv \tausyn \dot{I}+I=0
\end{align}
In the following, we will assume that all event times $\E$ are distinct, both in terms of spikes occurring and of spikes arriving. In continuous time, this is not unlikely, but also, as argued in \cite{wunderlich2021event}, the equations do not break down if spikes occur or arrive at the same time.
\begin{align}
    \frac{{\rm d}\mathcal{L}}{{\rm d}w_{ji}} &=\frac{{\rm d}}{{\rm d}w_{ji}}\left[l_p(\Ss)+\sum_{\tkev \in \E} \int_{\tkev}^{t_{k+1}^{\text{event}}}\left[l_V(V,t)+ \lambda_V \cdot f_V+\lambda_I \cdot f_I\right]\dt\right] \\
    \frac{\partial f_V}{\partial w_{ji}} &= \taum\frac{{\rm d}}{\dt}\frac{\partial V}{\partial w_{ji}}+\frac{\partial V}{\partial w_{ji}}-\frac{\partial I}{\partial w_{ji}} \label{eqn:dfv} \\
    \frac{\partial f_I}{\partial w_{ji}} &= \tausyn\frac{{\rm d}}{\dt}\frac{\partial I}{\partial w_{ji}}+\frac{\partial I}{\partial w_{ji}} \label{eqn:dfi}
\end{align}
We apply the derivative and use (\ref{eqn:dfv}) and (\ref{eqn:dfi}) to obtain
\begin{align}
    \frac{{\rm d}\mathcal{L}}{{\rm d}w_{ji}} &= \sum_{\tkspk \in \cal S} \frac{\partial l_p}{\partial \tkspk}\frac{{\rm d}\tkspk}{{\rm d}w_{ji}} \nonumber \\ 
&\hphantom{= } + \sum_{\tkev \in \E} \int_{\tkev}^{t_{k+1}^{\text{event}}} \left[\frac{\partial l_V}{\partial V} \cdot \frac{\partial V}{\partial w_{ji}}+\lambda_V \cdot \left(\taum\frac{{\rm d}}{\dt}\frac{\partial V}{\partial w_{ji}} + \frac{\partial V}{\partial w_{ji}} -\frac{\partial I}{\partial w_{ji}} \right ) \right. \label{eqn:dlw1} \\ 
 &\hphantom{= + \sum_{k=0}^{|\E|} \int_{\tkev}^{t_{k+1}^{\text{event}}} \Bigg[} 
\left. + \lambda_I \cdot \left(\tausyn\frac{{\rm d}}{\dt}\frac{\partial I}{\partial w_{ji}} + \frac{\partial I}{\partial w_{ji}}\right)\right]\dt  \nonumber \\
 &\hphantom{= + \sum_{k=0}^{|\E|}} + l_{V,k+1}^-\frac{\dt_{k+1}^{\text{event}}}{{\rm d}w_{ji}}  - l_{V,k}^+\frac{{\rm d}\tkev}{{\rm d}w_{ji}} \nonumber
\end{align}
Using partial integration, we obtain
\begin{align}
    \int_{\tkev}^{t_{k+1}^{\text{event}}} \lambda_V \cdot \frac{{\rm d}}{\dt} \frac{\partial V}{\partial w_{ji}} \dt &= - \int_{\tkev}^{t_{k+1}^{\text{event}}} \dot{\lambda}_V \cdot \frac{\partial V}{\partial w_{ji}} \dt + \left[\lambda_V \cdot \frac{\partial V}{\partial w_{ji}}\right]_{\tkev}^{t_{k+1}^{\text{event}}}
\end{align}
and
\begin{align}
    \int_{\tkev}^{t_{k+1}^{\text{event}}} \lambda_I \cdot \frac{{\rm d}}{\dt} \frac{\partial I}{\partial w_{ji}} \dt &= - \int_{\tkev}^{t_{k+1}^{\text{event}}} \dot{\lambda}_I \cdot \frac{\partial I}{\partial w_{ji}} \dt + \left[\lambda_I \cdot \frac{\partial I}{\partial w_{ji}}\right]_{\tkev}^{t_{k+1}^{\text{event}}}.
\end{align}
Inserting this into (\ref{eqn:dlw1}), we get
\begin{align}
    \frac{{\rm d}\mathcal{L}}{{\rm d}w_{ji}} &= \sum_{\tkspk \in \Ss} \frac{\partial l_p}{\partial \tkspk}\frac{{\rm d}\tkspk}{{\rm d}w_{ji}} \nonumber \\ 
   + \sum_{\tkev \in \E} &\int_{\tkev}^{t_{k+1}^{\text{event}}}\left[\left(\frac{\partial l_V}{\partial V}-\taum\dot{\lambda}_V+ \lambda_{V}\right) \cdot \frac{\partial V}{\partial w_{ji}} +(-\tausyn\dot{\lambda}_I+\lambda_{I}-\lambda_V) \cdot \frac{\partial I}{\partial w_{ji}}\right]\dt \nonumber \\
    &+\taum\left[\lambda_V \cdot \frac{\partial V}{\partial w_{ji}}\right]_{\tkev}^{t_{k+1}^{\text{event}}} + \tausyn\left[\lambda_I \cdot \frac{\partial I}{\partial w_{ji}}\right]_{\tkev}^{t_{k+1}^{\text{event}}} \\
    &+ l_{V,k+1}^-\frac{\dt_{k+1}^{\text{event}}}{{\rm d}w_{ji}}  - l_{V,k}^+\frac{{\rm d}\tkev} {{\rm d}w_{ji}} \nonumber
\end{align}
where the last two terms arise from the derivative of the bounds of the integral in the Leibniz rule.
We now define the backwards dynamics of the adjoint variables as before,
\begin{align}
    \taum \lambda_V^{'} &=-\lambda_V-\frac{\partial l_V}{\partial V} \\
    \tausyn \lambda_I^{'} &=-\lambda_I+\lambda_V
\end{align}
which cancels the terms containing $\frac{\partial V}{\partial w_{ji}}$ and $\frac{\partial I}{\partial w_{ji}}$, and we get
\begin{align}
    \frac{{\rm d}\mathcal{L}}{{\rm d}w_{ji}} &=  \sum_{\tkspk \in \cal S} \frac{\partial l_p}{\partial \tkspk}\frac{{\rm d}\tkspk}{{\rm d}w_{ji}} 
   + \sum_{\tkev \in \E}\Bigg(
        l_{V,k}^-\frac{{\rm d}\tkev}{{\rm d}w_{ji}} - l_{V,k}^+\frac{{\rm d}\tkev}{{\rm d}w_{ji}}  \label{eqn:dlsum} \\
        &+\left.\left[\taum\left(\lambda_V^- \cdot \frac{\partial V^{-}}{\partial w_{ji}}-\lambda_V^+ \cdot \frac{\partial V^{+}}{\partial w_{ji}}\right)+\tausyn\left(\lambda_I^- \cdot \frac{\partial I^{-}}{\partial w_{ji}}-\lambda_I^+ \cdot \frac{\partial I^{+}}{\partial w_{ji}}\right)\right]\right|_{\tkev}\Bigg) \nonumber
\end{align}
We now focus on the spike emission times $\tkspk \in \Ss$. Before the jump at $\tkspk$ we have,
\begin{equation}
    (V^-)_{n(k)}-\vartheta=0,
\end{equation}
where $n(k)$ denotes the spiking neuron at event $k$. If we take the derivative of this equation, we get, using the chain rule,
\begin{align}
    \left(\frac{\partial V^-}{\partial w_{ji}}\right)_{n(k)}+(\dot{V}^-)_{n(k)}\frac{{\rm d}\tkspk}{{\rm d}w_{ji}} &=0 \label{eqn:dvmdwji} \\
    \Rightarrow \quad \frac{{\rm d}\tkspk}{{\rm d}w_{ji}} &=-\frac{1}{(\dot{V}^-)_{n(k)}}\left(\frac{\partial V^-}{\partial w_{ji}}\right)_{n(k)}, \label{eqn:dtldwji}
\end{align}
and after the jump,
\begin{align}
    (V^+)_{n(k)} &=0 \\
    \Rightarrow \quad \left(\frac{\partial V^+}{\partial w_{ji}}\right)_{n(k)}+(\dot{V}^+)_{n(k)}\frac{{\rm d}\tkspk}{{\rm d}w_{ji}} &=0 \,.  \label{eqn:dvpdwji}
\end{align}
Inserting (\ref{eqn:dtldwji}) into (\ref{eqn:dvpdwji}) we obtain as usual
\begin{align}
    \left(\frac{\partial V^+}{\partial w_{ji}}\right)_{n(k)} &= \frac{(\dot{V}^+)_{n(k)}}{(\dot{V}^-)_{n(k)}}\left(\frac{\partial V^-}{\partial w_{ji}}\right)_{n(k)}. \label{eqn:dvpdwji2}
\end{align}
For the current $I_{n(k)}$, there is no jump at $\tkspk$, and also not in its derivative: $(I^+)_{n(k)} = (I^-)_{n(k)}$ and $(\dot{I}^+)_{n(k)} = (\dot{I}^-)_{n(k)}$ implies
\begin{equation}
    \left(\frac{\partial I^+}{\partial w_{ji}}\right)_{n(k)} = \left(\frac{\partial I^-}{\partial w_{ji}}\right)_{n(k)} .\label{eqn:dipdwji}
\end{equation}
Let us now consider what happens when the spike $k$ at $\tkspk$ is received at all the postsynaptic neurons $m$ at times $\tkspk + d_{mn(k)}$ (i.e. we look at $\E\setminus\Ss$). At these times, the input current of the receiving neurons jumps, 
\begin{equation}
    (I^+)_{m}=(I^-)_{m}+w_{mn(k)} \label{eqn:ijump} .
\end{equation}
By taking the derivative with respect to $w_{ji}$, we get
\begin{equation}
    \left(\frac{\partial I^+}{\partial w_{ji}}\right)_{m} + (\dot{I}^+)_{m} \frac{{\rm d}\tkspk}{{\rm d}w_{ji}} = \left(\frac{\partial I^-}{\partial w_{ji}}\right)_{m} + (\dot{I}^-)_{m} \frac{{\rm d}\tkspk}{{\rm d}w_{ji}} + \delta_{in(k)}\delta_{jm} ,
\end{equation}
where we have used that $\frac{{\rm d}(\tkspk +d_{mn(k)})}{{\rm d}w_{ji}} = \frac{{\rm d} \tkspk}{{\rm d}w_{ji}}$. Now, using the dynamics equations for $I$, we also have
\begin{equation}
    \tausyn(\dot{I}^+)_{m}=\tausyn(\dot{I}^-)_{m}-w_{mn(k)} ,
\end{equation}
and hence,
\begin{dmath}
    \left(\frac{\partial I^+}{\partial w_{ji}}\right)_{m} = \left(\frac{\partial I^-}{\partial w_{ji}}\right)_{m} + \tausyn^{-1}w_{mn(k)}\frac{{\rm d}\tkspk}{{\rm d}w_{ji}}+\delta_{in(k)}\delta_{jm} = \left(\frac{\partial I^-}{\partial w_{ji}}\right)_{m} + \left.\left[\frac{1}{\tausyn(\dot{V}^-)_{n(k)}}w_{mn(k)}\left(\frac{\partial V^-}{\partial w_{ji}}\right)_{n(k)}\right]\right|_{\tkspk +d_{mn(k)}}+\delta_{in(k)}\delta_{jm}
\end{dmath}
where we have used (\ref{eqn:dtldwji}) to replace $\frac{{\rm d}\tkspk}{{\rm d}w_{ji}}$.
Since we have $(V^+)_m = (V^-)_m$ for non-spiking neurons,
\begin{align}
\left(\frac{\partial V^+}{\partial w_{ji}}\right)_{m} + (\dot{V}^+)_{m} \frac{{\rm d}\tkspk}{{\rm d}w_{ji}} = \left(\frac{\partial V^-}{\partial w_{ji}}\right)_{m} + (\dot{V}^-)_{m} \frac{{\rm d}\tkspk}{{\rm d}w_{ji}}
\end{align}
From equation (\ref{eqn:ijump}) and the dynamics equations for $V$ we know
\begin{equation}
    \taum(\dot{V}^+)_{m}=\taum(\dot{V}^-)_{m}+w_{mn(k)} .
\end{equation}
%
Putting this together, we get 
\begin{align}
    \left(\frac{\partial V^+}{\partial w_{ji}}\right)_{m} &= \left(\frac{\partial V^-}{\partial w_{ji}}\right)_{m} -\taum^{-1}w_{mn(k)}\frac{{\rm d}\tkev}{{\rm d}w_{ji}} \\
    &= \left(\frac{\partial V^-}{\partial w_{ji}}\right)_{m} +\left.\left[\frac{1}{\taum(\dot{V}^-)_{n(k)}}w_{mn(k)}\left(\frac{\partial V^-}{\partial w_{ji}}\right)_{n(k)} \right]\right|_{\tkspk +d_{mn(k)}} \label{eqn:dvpdwjim}
\end{align}
%
%
We now can insert the expressions (\ref{eqn:dtldwji}), (\ref{eqn:dipdwji}), (\ref{eqn:dvpdwji2}) and (\ref{eqn:dvpdwjim}) into (\ref{eqn:dlsum}) and reorder terms according to which spike the jumps originate from, we get
\begin{align}
   \frac{{\rm d}\mathcal{L}}{{\rm d}w_{ji}}=   \sum_{\tkspk \in \cal S}   & \Bigg[ 
   \left(\frac{\partial V^-}{\partial w_{ji}}\right)_n\left[\taum\left(\lambda_V^--\frac{(\dot{V}^+)_{n}}{(\dot{V}^-)_{n}}\lambda_V^+\right)_n+\frac{1}{(\dot{V}^-)_{n}}\left(-\frac{\partial l_p}{\partial \tkspk}+ l_V^+-l_V^-\right)\right] \nonumber \\
   & +\tausyn(\lambda_I^--\lambda_I^+)_n\left(\frac{\partial I ^-}{\partial w_{ji}}\right)_n \Bigg]\Bigg|_{\tkspk}  \\
   &+ \sum_m \Bigg[\taum(\lambda_V^--\lambda_V^+)_{m}\left(\frac{\partial V^-}{\partial w_{ji}}\right)_m+\tausyn(\lambda_I^--\lambda_I^+)_m\left(\frac{\partial I^-}{\partial w_{ji}}\right)_m\Bigg]\Bigg|_{\tkspk+d_{mn(k)}} \nonumber \\
   &+ \left. \left[\left(\frac{\partial V^-}{\partial w_{ji}}\right)_n \frac{1}{(\dot{V}^-)_{n}}\right]\right|_{\tkspk}\left. \left[w_{mn}(\lambda_I^+-\lambda_V^+)_m\right]\right|_{\tkspk+d_{mn(k)}}-\left. \left[\tausyn\delta_{in(k)}\delta_{jm}(\lambda_I^+)_{m}\right]\right|_{\tkspk+d_{mn(k)}} . \nonumber
\end{align}
Thus, the update of $\lambda_V$ of the spiking neuron stays the same as without delays, apart from taking the receiving neurons' corresponding $\lambda_V$ and $\lambda_I$ at the delayed time.
\begin{align}
    (\lambda_V^-)_{n(k)} &= \left. \left[\frac{(\dot{V}^+)_{n(k)}}{(\dot{V}^-)_{n(k)}}(\lambda_V^+)_{n(k)}+\frac{1}{\taum(\dot{V}^-)_{n(k)}}\left[\frac{\partial l_p}{\partial \tkspk}+ l_V^--l_V^+\right]\right]\right|_{\tkspk} \label{eqn:delay_prop}\\
    & \hphantom{= } +\left. \left[\frac{1}{\taum(\dot{V}^-)_{n(k)}}\right]\right|_{\tkspk} \sum_{m}w_{mn(k)} \left.\left[(\lambda_V^+-\lambda_I^+)_m\right]\right|_{\tkspk+d_{mn(k)}} \nonumber \\
    (\lambda_V^-)_m &= (\lambda_V^+)_m \text{, if } m \neq n(k) \\
    \lambda_I^- &= \lambda_I^+ .
\end{align} 
The gradient is then given by
\begin{align}
    \frac{{\rm d}\mathcal{L}}{{\rm d}w_{ji}} &= -\tausyn \sum_{\tkspk \in \Ss}\delta_{in(k)}(\lambda_I)_j\Big|_{\tkspk+d_{jn(k)}}=-\tausyn \sum_{\parbox{0.9cm}{\scriptsize $\tkspk$ \parbox{0.9cm}{from\ $i$}}}(\lambda_I)_j\Big|_{\tkspk+d_{ji}} .
\end{align}
\subsubsection{Learning Delays}
In the following, we will derive the gradients for delays $d_{ji}$ similarly to our weight gradient derivations.
\begin{align}
    \frac{{\rm d}\mathcal{L}}{{\rm d}d_{ji}} &=\frac{{\rm d}}{{\rm d}d_{ji}}\left[l_p(\Ss)+\sum_{\tkev \in \E}\int_{\tkev}^{t_{k+1}^{\text{event}}}\left[l_V(V,t)+ \lambda_V \cdot f_V+\lambda_I \cdot f_I\right]\dt\right] \\
    \frac{\partial f_V}{\partial d_{ji}} &= \taum\frac{{\rm d}}{\dt}\frac{\partial V}{\partial d_{ji}}+\frac{\partial V}{\partial d_{ji}}-\frac{\partial I}{\partial d_{ji}} \\
    \frac{\partial f_I}{\partial d_{ji}} &= \tausyn\frac{{\rm d}}{\dt}\frac{\partial I}{\partial d_{ji}}+\frac{\partial I}{\partial d_{ji}}
\end{align}
Therefore,
\begin{align}
    \frac{{\rm d}\mathcal{L}}{{\rm d}d_{ji}} &= \sum_{\tkspk \in \cal S} \frac{\partial l_p}{\partial \tkspk}\frac{{\rm d}\tkspk}{{\rm d}d_{ji}} \nonumber \\ 
&\hphantom{= } + \sum_{\tkev \in \E}\int_{\tkev}^{t_{k+1}^{\text{event}}} \left[\frac{\partial l_V}{\partial V} \cdot \frac{\partial V}{\partial d_{ji}}+\lambda_V \cdot \left(\taum\frac{{\rm d}}{\dt}\frac{\partial V}{\partial d_{ji}} + \frac{\partial V}{\partial d_{ji}} -\frac{\partial I}{\partial d_{ji}} \right ) \right. \\ 
 &\hphantom{= + \sum_{k=0}^{|\E|} \int_{\tkev}^{t_{k+1}^{\text{event}}} \Bigg[} 
\left. + \lambda_I \cdot \left(\tausyn\frac{{\rm d}}{\dt}\frac{\partial I}{\partial d_{ji}} + \frac{\partial I}{\partial d_{ji}}\right)\right]\dt  \nonumber \\
 &\hphantom{= + \sum_{k=0}^{|\E|}} + l_{V,k+1}^-\frac{\dt_{k+1}^{\text{event}}}{{\rm d}d_{ji}}  - l_{V,k}^+\frac{{\rm d}\tkev}{{\rm d}d_{ji}}. \nonumber
\end{align}
Then, using partial integration,
\begin{align}
    \int_{\tkev}^{t_{k+1}^{\text{event}}} \lambda_V \cdot \frac{{\rm d}}{\dt} \frac{\partial V}{\partial d_{ji}} \dt &= - \int_{\tkev}^{t_{k+1}^{\text{event}}} \dot{\lambda}_V \cdot \frac{\partial V}{\partial d_{ji}} \dt + \left[\lambda_V \cdot \frac{\partial V}{\partial d_{ji}}\right]_{\tkev}^{t_{k+1}^{\text{event}}} \\
    \int_{\tkev}^{t_{k+1}^{\text{event}}} \lambda_I \cdot \frac{{\rm d}}{\dt} \frac{\partial I}{\partial d_{ji}} \dt &= - \int_{\tkev}^{t_{k+1}^{\text{event}}} \dot{\lambda}_I \cdot \frac{\partial I}{\partial d_{ji}} \dt + \left[\lambda_I \cdot \frac{\partial I}{\partial d_{ji}}\right]_{\tkev}^{t_{k+1}^{\text{event}}}
\end{align}
and hence,
\begin{align}
    \frac{{\rm d}\mathcal{L}}{{\rm d}d_{ji}} &= 
    \sum_{\tkspk \in S}  \frac{\partial l_p}{\partial \tkspk}\frac{{\rm d}\tkspk}{{\rm d}d_{ji}} \\
    &\sum_{\tkev \in \E}\left[\int_{\tkev}^{t_{k+1}^{\text{event}}}\left(\frac{\partial l_V}{\partial V}-\taum\dot{\lambda}_V+ \lambda_{V}\right) \cdot \frac{\partial V}{\partial d_{ji}}+(-\tausyn\dot{\lambda}_I+\lambda_{I}-\lambda_V) \cdot \frac{\partial I}{\partial d_{ji}}\right]\dt \nonumber \\
    &+\taum\left[\lambda_V \cdot \frac{\partial V}{\partial d_{ji}}\right]_{\tkev}^{t_{k+1}^{\text{event}}} + \tausyn\left[\lambda_I \cdot \frac{\partial I}{\partial d_{ji}}\right]_{\tkev}^{t_{k+1}^{\text{event}}}+ l_{V,k+1}^-\frac{\dt_{k+1}^{\text{event}}}{{\rm d}d_{ji}}  - l_{V,k}^+\frac{{\rm d}\tkev}{{\rm d}d_{ji}} . \nonumber
\end{align}
If we now define the adjoint dynamics as usual, the terms in the integral disappear, and we are left with
\begin{align}
 \frac{{\rm d}\mathcal{L}}{{\rm d}d_{ji}} &= 
  \sum_{\tkspk \in \Ss}  \frac{\partial l_p}{\partial \tkspk}\frac{{\rm d}\tkspk}{{\rm d}d_{ji}} \\
  &+ \sum_{\tkev \in \E} l_{V,k}^-\frac{{\rm d}\tkev}{{\rm d}d_{ji}} - l_{V,k}^+\frac{{\rm d}\tkev}{{\rm d}d_{ji}}+ \left. \left[\taum\left(\lambda_V^- \cdot \frac{\partial V^{-}}{\partial d_{ji}}-\lambda_V^+ \cdot \frac{\partial V^{+}}{\partial d_{ji}}\right)+\tausyn\left(\lambda_I^- \cdot \frac{\partial I^{-}}{\partial d_{ji}}-\lambda_I^+ \cdot \frac{\partial I^{+}}{\partial d_{ji}}\right)\right]\right|_{\tkev} . \nonumber
\end{align}
Let's now again first consider the spike emission times $\tkspk$ and the spiking neuron $n(k)$. Before the jump:
\begin{align}
    \left(\frac{\partial V^-}{\partial d_{ji}}\right)_{n(k)}+(\dot{V}^-)_{n(k)}\frac{{\rm d}\tkspk}{{\rm d}d_{ji}} &=0 \\
\Rightarrow \quad \frac{{\rm d}\tkspk}{{\rm d}d_{ji}} &=-\frac{1}{(\dot{V}^-)_{n(k)}}\left(\frac{\partial V^-}{\partial d_{ji}}\right)_{n(k)} , \label{eqn:dtlddji}
\end{align}
and after the jump:
\begin{align}
    \left(\frac{\partial V^+}{\partial d_{ji}}\right)_{n(k)}+(\dot{V}^+)_{n(k)}\frac{{\rm d}\tkspk}{{\rm d}d_{ji}} &=0 \\
   \Rightarrow \quad \left(\frac{\partial V^+}{\partial d_{ji}}\right)_{n(k)} &= \frac{(\dot{V}^+)_{n(k)}}{(\dot{V}^-)_{n(k)}}\left(\frac{\partial V^-}{\partial d_{ji}}\right)_{n(k)}.
\end{align}
There is no jump in $I_{n(k)}$ or its time derivative at $\tkspk$ which analogous to above implies
\begin{align}
        \left(\frac{\partial I^+}{\partial d_{ji}}\right)_{n(k)} = \left(\frac{\partial I^-}{\partial d_{ji}}\right)_{n(k)} .
\end{align}
Turning to spike arrival times $\tkev \in \E \backslash \Ss$, when the spike at $\tkspk$ arrives at the post-synaptic neurons $m$, we get 
\begin{align}
    (I^+)_{m}=(I^-)_{m}+w_{mn(k)}, \label{eqn:ijumpd}
\end{align}
and hence,
\begin{align}
    \left(\frac{\partial I^+}{\partial d_{ji}}\right)_{m} + (\dot{I}^+)_{m} \frac{{\rm d}\tkev}{{\rm d}d_{ji}} = \left(\frac{\partial I^-}{\partial d_{ji}}\right)_{m} + (\dot{I}^-)_{m} \frac{{\rm d}\tkev}{{\rm d}d_{ji}} .
\end{align}
Using the dynamics of $I$, (\ref{eqn:ijumpd}) implies
\begin{equation}
    \tausyn(\dot{I}^+)_{m(l,n)}=\tausyn(\dot{I}^-)_{m(l,n)}-w_{mn} ,
\end{equation}
and hence
\begin{align}
    \left(\frac{\partial I^+}{\partial d_{ji}}\right)_{m} &= \left(\frac{\partial I^-}{\partial d_{ji}}\right)_{m} + \tausyn^{-1}w_{mn(k)}\frac{{\rm d}\tkev}{{\rm d}d_{ji}} \\
    &= \left(\frac{\partial I^-}{\partial d_{ji}}\right)_{m} - \frac{1}{\tausyn(\dot{V}^-)_{n(k)}}w_{mn(k)}\left(\frac{\partial V^-}{\partial d_{ji}}\right)_{n(k)}+\delta_{in(k)}\delta_{jm} \frac{w_{mn(k)}}{{\tausyn}},
\end{align}
where the term involving the spiking neuron $n(k)$ stems from the derivative of the spike time $\tkev$ with respect to $d_{ji}$ using (\ref{eqn:dtlddji}) and the last term from the derivative of the delay by itself. For the voltages, 
\begin{align}
    \left(\frac{\partial V^+}{\partial d_{ji}}\right)_{m} + (\dot{V}^+)_{m} \frac{{\rm d}\tkev}{{\rm d}d_{ji}} = \left(\frac{\partial V^-}{\partial d_{ji}}\right)_{m} + (\dot{V}^-)_{m} \frac{{\rm d}\tkev}{{\rm d}d_{ji}} ,
\end{align}
and using the dynamics of $V$ and (\ref{eqn:ijumpd}),
\begin{equation}
    \taum(\dot{V}^+)_{m}=\taum(\dot{V}^-)_{m}+w_{mn(k)} ,
\end{equation}
which put together gives
\begin{align}
    \left(\frac{\partial V^+}{\partial d_{ji}}\right)_{m} &= \left(\frac{\partial V^-}{\partial d_{ji}}\right)_{m} -\taum^{-1}w_{mn}\frac{{\rm d}\tkev}{{\rm d}d_{ji}} \\
    &= \left(\frac{\partial V^-}{\partial d_{ji}}\right)_{m} +\frac{1}{\taum(\dot{V}^-)_{n(k)}}w_{mn(k)}\left(\frac{\partial V^-}{\partial d_{ji}}\right)_{n(k)}-\delta_{in(k)}\delta_{jm} \frac{w_{mn(k)}}{{\taum}} ,
\end{align}
where the last term again arises from the derivative of the delay $d_{mn(k)}$ in $\tkev$ with respect to $d_{ji}$. 
Taking everything together we get
\begin{align}
    \frac{d {\cal L}}{d d_{ji}} &= \sum_{\tkspk \in \Ss}  \left[
    \left(\frac{\partial V^-}{\partial d_{ji}}\right)_n\left[\taum\left(\lambda_V^--\frac{(\dot{V}^+)_{n}}{(\dot{V}^-)_{n}}\lambda_V^+\right)_n+\frac{1}{(\dot{V}^-)_{n}}\left(-\frac{\partial l_p}{\partial \tkspk}+ l_V^+-l_V^-\right)\right] \right.\\
    &\left. \left. +\tausyn(\lambda_I^--\lambda_I^+)_n\left(\frac{\partial I ^-}{\partial d_{ji}}\right)_n\right]\right|_{\tkspk} \\
    &+ \sum_m \left.\left[ \taum(\lambda_V^--\lambda_V^+)_{m}\left(\frac{\partial V^-}{\partial d_{ji}}\right)_m+\tausyn(\lambda_I^--\lambda_I^+)_m\left(\frac{\partial I^-}{\partial d_{ji}}\right)_m\right]\right|_{\tkspk +d_{mn(k)}} \\
    &+ \left. \left[\left(\frac{\partial V^-}{\partial d_{ji}}\right)_n \frac{1}{(\dot{V}^-)_{n}}\right]\right|_{\tkspk}\left. \left[w_{mn}(\lambda_I^+-\lambda_V^+)_m\right]\right|_{\tkspk+d_{mn(k)}} \\
    &-\left. \left[w_{mn(k)}\delta_{in(k)}\delta_{jm}(\lambda_I^+-\lambda_V^+)_{m}\right]\right|_{\tkspk+d_{mn(k)}} .
\end{align}
So, using the usual trick
\begin{align}
    \frac{(\dot{V}^+)_{n(k)}}{(\dot{V}^-)_{n(k)}} = \frac{\vartheta}{\taum(\dot{V}^-)_{n(k)}}+1, 
\end{align} we again arrive at the same jump conditions as usual,
\begin{align}
    (\lambda_V^-)_{n(k)} &= \left. \left[ (\lambda_V^+)_{n(k)} +\frac{1}{\taum(\dot{V}^-)_{n(k)}}\left[\vartheta \cdot (\lambda_V^+)_{n(k)} +\frac{\partial l_p}{\partial \tkspk}+ l_V^--l_V^+\right]\right]\right|_{\tkspk} \\
    & \hphantom{= } +\left. \left[\frac{1}{\taum(\dot{V}^-)_{n(k)}}\right]\right|_{\tkspk} \sum_{m}w_{mn(k)} \left.\left[(\lambda_V^+-\lambda_I^+)_m\right]\right|_{\tkspk+d_{mn(k)}} \nonumber \\
    (\lambda_V^-)_m &= (\lambda_V^+)_m \text{, if } m \neq n(k) \\
    \lambda_I^- &= \lambda_I^+ ,
\end{align}
but the gradient updates take the form
\begin{equation}
    \frac{{\rm d}\mathcal{L}}{{\rm d}d_{ji}} = -\sum_{\tkspk \in \Ss}w_{ji}\delta_{in(k)}\left.(\lambda_I-\lambda_V)_j\right|_{\tkspk+d_{jn(k)}}= -w_{ji}\sum_{\parbox{0.9cm}{\scriptsize $\tkspk $ \parbox{0.9cm}{ from\ $i$}}} \left.(\lambda_I-\lambda_V)_j\right|_{\tkspk+d_{ji}} .
\end{equation}

\subsubsection{Time-invariant mean squared error loss}
Following \cite{goltz2024delgrad}, we use for the Yin-yang benchmark the time-invariant mean squared error loss of output spike times
\begin{align}
    {\cal L}_{\Delta \rm MSE} = \frac{1}{2} \sum_{i \neq c} \left( t_i - t_{c} - \Delta_t \right)^2 ,
\end{align}
where $c$ is the true class of the current input and $t_i$, $t_c$ denote the first spike time in the respective output neurons. 
In the EventProp formalism, this is a spike-time dependent loss $l_p$ and, therefore, drives jumps in $\lambda_V^i$ in output neuron $i$ at spike times $\tkspk$ in the backward pass (see Table \ref{tab:dynamics}) by
\begin{align}
\frac{\partial l_p}{\partial \tkspk} = \left\{ \begin{array}{ll} (t_i - t_c - \Delta_t) & \text{ if } n(k)=i, \tkspk=t_i, i \neq c \\
\sum_{i \neq c} - (t_i - t_c - \Delta_t) & \text{ if } n(k)=c, \tkspk=t_c \\
0 & \text{otherwise}
\end{array}
\right.
\end{align}

\begin{table}
  \caption{Yin-yang parameters}
  \label{tab:yyparam}
  \begin{tabular}{ll}
    \toprule
    Architecture  &   Feedforward \\
    \midrule
    Number of hidden layers & $1$ \\
    Number of hidden neurons & $30/25/20/15/10/5$\\
    Input-to-hidden weight init.  & $\mathcal{N}(2.0,\,0.78)$\\
    Hidden-to-out weight init. & $\mathcal{N}(0.93,\,0.1)$\\
    Ff delay init. & $\mathcal{U}(0,0)$\\
    DT & $0.01$\\
    Weight LR & $0.001$\\
    Weight LR schedule & $0.998 \times epoch$  \\
    Delay LR init. & $0.01$\\
    Delay LR schedule & $0.998 \times epoch$\\
    
    \bottomrule
  \end{tabular}
\end{table}

\begin{table}
  \caption{SHD parameters}
  \label{tab:shdparam}
  \begin{tabular}{lll}
    \toprule
    Architecture &  Recurrent & Feedforward\\
    \midrule
    Number of hidden layers & $1$ & $2$ \\
    Number of hidden neurons & $1024/512/256/128$ & $1024/512/256/128$ \\
    Spike reg. strength (layer-wise) & $5\cdot 10^{-11}$ & $5\cdot 10^{-12},5\cdot 10^{-11}$ \\
    Input-to-hidden weight init. & $\mathcal{N}(0.03,\,0.01)$ & $\mathcal{N}(0.03,\,0.01)$\\
    Ff hidden-to-hidden weight init. & \xmark & $\mathcal{N}(0.02,\,0.03)$ \\
    Rec hidden-to-hidden weight init. & $\mathcal{N}(0.0,\,0.02)$ & \xmark\\
    Hidden-to-out weight init. & $\mathcal{N}(0.02,\,0.03)$ & $\mathcal{N}(0.02,\,0.03)$\\
    Ff delay init. & $\mathcal{U}(0,150)$ & $\mathcal{U}(0,100)$\\
    Rec delay init. & $\mathcal{U}(0,0)$& \xmark\\
    DT & $1.0$& $1.0$\\
    Weight LR & $0.001$& $0.001$\\
    Weight LR schedule & $1.05 ^ {batch}$ & $1.05 ^ {batch}$  \\
    Delay LR init. & $0.1$& $0.1$\\
    Delay LR schedule & \xmark & \xmark \\
    
    \bottomrule
  \end{tabular}
\end{table}

\begin{table}
  \caption{SSC parameters}
  \label{tab:sscparam}
  \begin{tabular}{lll}
    \toprule
    Architecture  & Recurrent &  Feedforward \\
    \midrule
    Number of hidden layers & $1$  & $2$ \\
    Number of hidden neurons & $1024/512/256/128/64$  & $1024/512/256/128/64$ \\
    Spike reg. strength (layer-wise) & $5\cdot 10^{-11}$ & $5\cdot 10^{-12},5\cdot 10^{-12}$ \\
    Input-to-hidden weight init.  & $\mathcal{N}(0.03,\,0.01)$ & $\mathcal{N}(0.03,\,0.01)$ \\
    Ff hidden-to-hidden weight init. & \xmark & $\mathcal{N}(0.02,\,0.03)$\\
    Rec hidden-to-hidden weight init. &  $\mathcal{N}(0.0,\,0.02)$ & \xmark \\
    Hidden-to-out weight init. & $\mathcal{N}(0.02,\,0.3)$ & $\mathcal{N}(0.02,\,0.03)$ \\
    Ff delay init. & $\mathcal{U}(0,150)$ & $\mathcal{U}(0,50)$ \\
    Rec delay init.  & $\mathcal{U}(0,0)$ & \xmark \\
    DT & $1.0$ & $1.0$\\
    Weight LR & $0.001$ & $0.001$\\
    Weight LR schedule & $1.05 ^ {batch}$ &  $1.05 ^ {batch}$ \\
    Delay LR init. & $0.1$ & $0.1$\\
    Delay LR schedule & \xmark & \xmark\\
    
    \bottomrule
  \end{tabular}
\end{table}

\begin{table}
  \caption{Braille reading parameters}
  \label{tab:brailleparam}
  \begin{tabular}{lll}
    \toprule
    Architecture  & Recurrent & Feedforward\\
    \midrule
    Number of hidden layers & $1$ & $2$ \\
    Number of hidden neurons & $1024/512/256/128/64$  & $1024/512/256/128/64$ \\
    Spike reg. strength (layer-wise) & $1\cdot 10^{-10}$ & $1\cdot 10^{-10},1\cdot 10^{-10}$\\
    Input-to-hidden weight init.  &  $\mathcal{N}(0.03,\,0.01)$ & $\mathcal{N}(0.03,\,0.01)$\\
    Ff hidden-to-hidden weight init. & \xmark  & $\mathcal{N}(0.02,\,0.03)$\\
    Rec hidden-to-hidden weight init. & $\mathcal{N}(0.0,\,0.02)$ & \xmark\\
    Hidden-to-out weight init. & $\mathcal{N}(0.02,\,0.03)$ & $\mathcal{N}(0.02,\,0.03)$\\
    Ff delay init. & $\mathcal{U}(0,0)$ & $\mathcal{U}(0,0)$\\
    Rec delay init. & \xmark & \xmark\\
    DT & $4.0$ & $4.0$\\
    Weight LR & $0.0015$ & $0.0015$\\
    Weight LR schedule  & \xmark & \xmark \\
    Delay LR init. & $0.025$ & $0.025$\\
    Delay LR schedule & \xmark & \xmark \\
    
    \bottomrule
  \end{tabular}
\end{table}
\subsection{Implementation}
We implemented all of our work in the mlGeNN framework~\cite{knight2023easy,Turner2022} to exploit the the efficiency of event-based learning. In all of our experiments, we used the parameters from previous EventProp work~\cite{nowotny2022loss}, apart from spike regularisation strengths, number of hidden layers and recurrent connections. We did not implement heterogeneous and trainable time constants, so that the independent effect of delays would be more clear. 
For our experiments on the SHD and SSC datasets, we adopted the data augmentation approaches described by \citet{nowotny2022loss}, which were designed to improve generalization. Specifically, we implemented the following augmentations:
\begin{itemize}
    \item Input Shifting: We randomly shifted all inputs by a value within the range of (-40,40).
    \item Input Blending: We blended two inputs from the same class by aligning their centers of mass and randomly selecting spikes from each input with a probability of 0.5.
\end{itemize}
For SSC we only used the shift augmentation. For the Yin-Yang dataset we decreased the learning rate on both weights and delays at the end of each epoch. On SHD and SSC, we implemented an ``ease-in'' scheduler on the weight learning rate, starting from $0.001$ times the learning rate, increasing it at the end of each batch, until it reached the final value. For our chosen hyperparameters, see Tables \ref{tab:yyparam},  \ref{tab:shdparam}, \ref{tab:sscparam} and \ref{tab:brailleparam}. 
GeNN already provided an efficient implementation of spike transmission with per-synapse delays~\citep{Knight2018} -- allowing the EventProp forward pass to be implemented efficiently.
However, the $\lambda_V$ transitions in the backward pass require access to postsynaptic $\lambda$ values with a per-synapse delay ($\left[(\lambda_V^+ -\lambda_I^+)_m\right]|_{\tkspk+d_{mn(k)}}$ from Equation:~\ref{eqn:delay_prop}).
This required a small extension to GeNN's existing system for providing delayed access to postsynaptic variables from a synapse model~\citep{Yavuz2016} in order to enable it to use the per-synapse delays used for spike transmission in the forward pass.

\section*{Code availability}
All experiments were carried out using the GeNN 5.1.0~\cite{genn_5_1_0} an mlGeNN 2.3.0~\cite{ml_genn_2_3_0}. 
The latest versions of both libraries are also available at \url{https://github.com/genn-team/}.
The code to train and evaluate the models described in this work are available at \url{https://github.com/mbalazs98/deventprop} 

\section*{Data availability}
The data underlying our results are available at \url{https://figshare.com/s/e4a041d66c93355f586a}.

\section*{Acknowledgments}
This work was funded by the EPSRC (grants EP/V052241/1 and EP/S030964/1) and the EU (grant no. 945539). Additionally, BM was funded by a Leverhulme Trust studentship and by The Alan Turing Institute.
Compute time was provided through Gauss Centre for Supercomputing (application numbers 21018, 30182 and 61883) and EPSRC (grant number EP/T022205/1) and local GPU hardware was provided by an NVIDIA hardware grant award.
\printbibliography

@article{bohte_error-backpropagation_2002,
	title = {Error-backpropagation in temporally encoded networks of spiking neurons},
	volume = {48},
	issn = {09252312},
	url = {https://linkinghub.elsevier.com/retrieve/pii/S0925231201006580},
	doi = {10.1016/S0925-2312(01)00658-0},
	language = {en},
	number = {1-4},
	urldate = {2023-09-08},
	journal = {Neurocomputing},
	author = {Bohte, Sander M. and Kok, Joost N. and La Poutré, Han},
	month = oct,
	year = {2002},
	pages = {17--37},
}

@article{stanojevic_high-performance_2024,
	title = {High-performance deep spiking neural networks with 0.3 spikes per neuron},
	volume = {15},
	issn = {2041-1723},
	url = {https://www.nature.com/articles/s41467-024-51110-5},
	doi = {10.1038/s41467-024-51110-5},
	language = {en},
	number = {1},
	urldate = {2024-08-19},
	journal = {Nature Communications},
	author = {Stanojevic, Ana and Woźniak, Stanisław and Bellec, Guillaume and Cherubini, Giovanni and Pantazi, Angeliki and Gerstner, Wulfram},
	month = aug,
	year = {2024},
	pages = {6793},
}

@article{patterson2021carbon,
  title={Carbon emissions and large neural network training},
  author={Patterson, David and Gonzalez, Joseph and Le, Quoc and Liang, Chen and Munguia, Lluis-Miquel and Rothchild, Daniel and So, David and Texier, Maud and Dean, Jeff},
  journal={arXiv preprint arXiv:2104.10350},
  year={2021}
}

@article{stockl2021optimized,
  title={Optimized spiking neurons can classify images with high accuracy through temporal coding with two spikes},
  author={St{\"o}ckl, Christoph and Maass, Wolfgang},
  journal={Nature Machine Intelligence},
  volume={3},
  number={3},
  pages={230--238},
  year={2021},
  publisher={Nature Publishing Group UK London}
}

@article{sokoloff1992brain,
  title={The brain as a chemical machine},
  author={Sokoloff, Louis},
  journal={Progress in brain research},
  volume={94},
  pages={19--33},
  year={1992},
  publisher={Elsevier}
}

@article{sengupta2019going,
  title={Going deeper in spiking neural networks: VGG and residual architectures},
  author={Sengupta, Abhronil and Ye, Yuting and Wang, Robert and Liu, Chiao and Roy, Kaushik},
  journal={Frontiers in neuroscience},
  volume={13},
  pages={95},
  year={2019},
  publisher={Frontiers Media SA}
}

@article{zhuspikegpt,
  title={Spike{GPT}: Generative Pre-trained Language Model with Spiking Neural Networks},
author={Rui-Jie Zhu and Qihang Zhao and Guoqi Li and Jason Eshraghian},
journal={Transactions on Machine Learning Research},
issn={2835-8856},
year={2024},
url={https://openreview.net/forum?id=gcf1anBL9e},
}

@article{rueckauer2017conversion,
  title={Conversion of continuous-valued deep networks to efficient event-driven networks for image classification},
  author={Rueckauer, Bodo and Lungu, Iulia-Alexandra and Hu, Yuhuang and Pfeiffer, Michael and Liu, Shih-Chii},
  journal={Frontiers in neuroscience},
  volume={11},
  pages={682},
  year={2017},
  publisher={Frontiers Media SA}
}

@article{neftci2019surrogate,
  title={Surrogate gradient learning in spiking neural networks: Bringing the power of gradient-based optimization to spiking neural networks},
  author={Neftci, Emre O and Mostafa, Hesham and Zenke, Friedemann},
  journal={IEEE Signal Processing Magazine},
  volume={36},
  number={6},
  pages={51--63},
  year={2019},
  publisher={IEEE}
}

@article{
spikingjelly,
author = {Wei Fang  and Yanqi Chen  and Jianhao Ding  and Zhaofei Yu  and Timothée Masquelier  and Ding Chen  and Liwei Huang  and Huihui Zhou  and Guoqi Li  and Yonghong Tian },
title = {SpikingJelly: An open-source machine learning infrastructure platform for spike-based intelligence},
journal = {Science Advances},
volume = {9},
number = {40},
pages = {eadi1480},
year = {2023},
doi = {10.1126/sciadv.adi1480},
URL = {https://www.science.org/doi/abs/10.1126/sciadv.adi1480},
eprint = {https://www.science.org/doi/pdf/10.1126/sciadv.adi1480},
}

@misc{pehle_event-based_2023,
	title = {Event-based {Backpropagation} for {Analog} {Neuromorphic} {Hardware}},
	url = {http://arxiv.org/abs/2302.07141},
	doi = {10.48550/arXiv.2302.07141},
	urldate = {2024-12-17},
	publisher = {arXiv},
	author = {Pehle, Christian and Blessing, Luca and Arnold, Elias and Müller, Eric and Schemmel, Johannes},
	month = feb,
	year = {2023},
	note = {arXiv:2302.07141 [q-bio]},
}

@article{wunderlich2021event,
  title={Event-based backpropagation can compute exact gradients for spiking neural networks},
  author={Wunderlich, Timo C and Pehle, Christian},
  journal={Scientific Reports},
  volume={11},
  number={1},
  pages={12829},
  year={2021},
  publisher={Nature Publishing Group UK London}
}

@article{lecun1998gradient,
  title={Gradient-based learning applied to document recognition},
  author={LeCun, Yann and Bottou, L{\'e}on and Bengio, Yoshua and Haffner, Patrick},
  journal={Proceedings of the IEEE},
  volume={86},
  number={11},
  pages={2278--2324},
  year={1998},
  publisher={Ieee}
}

@inproceedings{kriener2022yin,
  title={The Yin-Yang dataset},
  author={Kriener, Laura and G{\"o}ltz, Julian and Petrovici, Mihai A},
  booktitle={Proceedings of the 2022 Annual Neuro-Inspired Computational Elements Conference},
  pages={107--111},
  year={2022}
}

@article{cramer2020heidelberg,
  title={The Heidelberg spiking data sets for the systematic evaluation of spiking neural networks},
  author={Cramer, Benjamin and Stradmann, Yannik and Schemmel, Johannes and Zenke, Friedemann},
  journal={IEEE Transactions on Neural Networks and Learning Systems},
  volume={33},
  number={7},
  pages={2744--2757},
  year={2020},
  publisher={IEEE}
}

@article{nowotny2022loss,
	author={Nowotny, Thomas and Turner, James Paul and Knight, James Courtney},
	title={Loss shaping enhances exact gradient learning with Eventprop in Spiking Neural Networks},
	journal={Neuromorphic Computing and Engineering},
	url={http://iopscience.iop.org/article/10.1088/2634-4386/ada852},
	year={2025},
}

@inproceedings{fang2021incorporating,
  title={Incorporating learnable membrane time constant to enhance learning of spiking neural networks},
  author={Fang, Wei and Yu, Zhaofei and Chen, Yanqi and Masquelier, Timoth{\'e}e and Huang, Tiejun and Tian, Yonghong},
  booktitle={Proceedings of the IEEE/CVF international conference on computer vision},
  pages={2661--2671},
  year={2021}
}

@inproceedings{hammouamrilearning,
  title={Learning Delays in Spiking Neural Networks using Dilated Convolutions with Learnable Spacings},
  author={Hammouamri, Ilyass and Khalfaoui-Hassani, Ismail and Masquelier, Timoth{\'e}e},
  booktitle={The Twelfth International Conference on Learning Representations}
}

@article{bengtsson2005extensive,
  title={Extensive piano practicing has regionally specific effects on white matter development},
  author={Bengtsson, Sara L and Nagy, Zolt{\'a}n and Skare, Stefan and Forsman, Lea and Forssberg, Hans and Ull{\'e}n, Fredrik},
  journal={Nature neuroscience},
  volume={8},
  number={9},
  pages={1148--1150},
  year={2005},
  publisher={Nature Publishing Group US New York}
}

@article{seidl2010mechanisms,
  title={Mechanisms for adjusting interaural time differences to achieve binaural coincidence detection},
  author={Seidl, Armin H and Rubel, Edwin W and Harris, David M},
  journal={Journal of Neuroscience},
  volume={30},
  number={1},
  pages={70--80},
  year={2010},
  publisher={Soc Neuroscience}
}

@article{izhikevich2006polychronization,
  title={Polychronization: computation with spikes},
  author={Izhikevich, Eugene M},
  journal={Neural computation},
  volume={18},
  number={2},
  pages={245--282},
  year={2006},
  publisher={MIT Press One Rogers Street, Cambridge, MA 02142-1209, USA journals-info~…}
}

@article{maass1999complexity,
  title={On the complexity of learning for spiking neurons with temporal coding},
  author={Maass, Wolfgang and Schmitt, Michael},
  journal={Information and Computation},
  volume={153},
  number={1},
  pages={26--46},
  year={1999},
  publisher={Elsevier}
}

@article{furber2014spinnaker,
  title={The SpiNNaker project},
  author={Furber, Steve B and Galluppi, Francesco and Temple, Steve and Plana, Luis A},
  journal={Proceedings of the IEEE},
  volume={102},
  number={5},
  pages={652--665},
  year={2014},
  publisher={IEEE}
}

@article{davies2018loihi,
  title={Loihi: A neuromorphic manycore processor with on-chip learning},
  author={Davies, Mike and Srinivasa, Narayan and Lin, Tsung-Han and Chinya, Gautham and Cao, Yongqiang and Choday, Sri Harsha and Dimou, Georgios and Joshi, Prasad and Imam, Nabil and Jain, Shweta and others},
  journal={Ieee Micro},
  volume={38},
  number={1},
  pages={82--99},
  year={2018},
  publisher={IEEE}
}

@article{sun2023learnable,
  title={Learnable axonal delay in spiking neural networks improves spoken word recognition},
  author={Sun, Pengfei and Chua, Yansong and Devos, Paul and Botteldooren, Dick},
  journal={Frontiers in Neuroscience},
  volume={17},
  pages={1275944},
  year={2023},
  publisher={Frontiers Media SA}
}

@article{deckers2024co,
  title={Co-learning synaptic delays, weights and adaptation in spiking neural networks},
  author={Deckers, Lucas and Van Damme, Laurens and Van Leekwijck, Werner and Tsang, Ing Jyh and Latr{\'e}, Steven},
  journal={Frontiers in Neuroscience},
  volume={18},
  pages={1360300},
  year={2024},
  publisher={Frontiers Media SA}
}

@article{goltz2024delgrad,
  title={DelGrad: Exact gradients in spiking networks for learning transmission delays and weights},
  author={G{\"o}ltz, Julian and Weber, Jimmy and Kriener, Laura and Lake, Peter and Payvand, Melika and Petrovici, Mihai A},
  journal={arXiv preprint arXiv:2404.19165},
  year={2024}
}

@inproceedings{schone2024scalable,
  title={Scalable event-by-event processing of neuromorphic sensory signals with deep state-space models},
  author={Sch{\"o}ne, Mark and Sushma, Neeraj Mohan and Zhuge, Jingyue and Mayr, Christian and Subramoney, Anand and Kappel, David},
  booktitle={2024 International Conference on Neuromorphic Systems (ICONS)},
  pages={124--131},
  year={2024},
  organization={IEEE}
}

@article{goltz2021fast,
  title={Fast and energy-efficient neuromorphic deep learning with first-spike times},
  author={G{\"o}ltz, Julian and Kriener, Laura and Baumbach, Andreas and Billaudelle, Sebastian and Breitwieser, Oliver and Cramer, Benjamin and Dold, Dominik and Kungl, Akos Ferenc and Senn, Walter and Schemmel, Johannes and others},
  journal={Nature machine intelligence},
  volume={3},
  number={9},
  pages={823--835},
  year={2021},
  publisher={Nature Publishing Group UK London}
}

@inproceedings{knight2023easy,
  title={Easy and efficient spike-based Machine Learning with mlGeNN},
  author={Knight, James C and Nowotny, Thomas},
  booktitle={Proceedings of the 2023 Annual Neuro-Inspired Computational Elements Conference},
  pages={115--120},
  year={2023}
}

@article{barton2002modeling,
  title={Modeling, simulation, sensitivity analysis, and optimization of hybrid systems},
  author={Barton, Paul I and Lee, Cha Kun},
  journal={ACM Transactions on Modeling and Computer Simulation (TOMACS)},
  volume={12},
  number={4},
  pages={256--289},
  year={2002},
  publisher={ACM New York, NY, USA}
}

@article{bittar2022surrogate,
  title={A surrogate gradient spiking baseline for speech command recognition},
  author={Bittar, Alexandre and Garner, Philip N},
  journal={Frontiers in Neuroscience},
  volume={16},
  pages={865897},
  year={2022},
  publisher={Frontiers Media SA}
}

@article{meszaros18learning,
  title={Learning Delays Through Gradients and Structure: Emergence of Spatiotemporal Patterns in Spiking Neural Networks},
  author={M{\'e}sz{\'a}ros, Bal{\'a}zs and Knight, James C and Nowotny, Thomas},
  journal={Frontiers in Computational Neuroscience},
  volume={18},
  pages={1460309},
  publisher={Frontiers}
}

@article{swadlow1985physiological,
  title={Physiological properties of individual cerebral axons studied in vivo for as long as one year},
  author={Swadlow, Harvey A},
  journal={Journal of neurophysiology},
  volume={54},
  number={5},
  pages={1346--1362},
  year={1985},
  publisher={American Physiological Society Bethesda, MD}
}

@article{moro2024role,
  title={The Role of Temporal Hierarchy in Spiking Neural Networks},
  author={Moro, Filippo and Aceituno, Pau Vilimelis and Kriener, Laura and Payvand, Melika},
  journal={arXiv preprint arXiv:2407.18838},
  year={2024}
}

@article{baronig2024advancing,
  title={Advancing Spatio-Temporal Processing in Spiking Neural Networks through Adaptation},
  author={Baronig, Maximilian and Ferrand, Romain and Sabathiel, Silvester and Legenstein, Robert},
  journal={arXiv preprint arXiv:2408.07517},
  year={2024}
}

@article{perez2021neural,
  title={Neural heterogeneity promotes robust learning},
  author={Perez-Nieves, Nicolas and Leung, Vincent CH and Dragotti, Pier Luigi and Goodman, Dan FM},
  journal={Nature communications},
  volume={12},
  number={1},
  pages={5791},
  year={2021},
  publisher={Nature Publishing Group UK London}
}

@article{selvaratnam2000learning,
  title={Learning methods of recurrent spiking neural networks-transient and oscillatory spike trains},
  author={Selvaratnam, Kukan},
  journal={Systems, control and information},
  volume={13},
  number={3},
  pages={95--104},
  year={2000}
}

@article{Knight2018,
	title = {{GPUs} Outperform Current {HPC} and Neuromorphic Solutions in Terms of Speed and Energy When Simulating a Highly-Connected Cortical Model},
	volume = {12},
	rights = {All rights reserved},
	issn = {1662-453X},
	url = {https://www.frontiersin.org/article/10.3389/fnins.2018.00941/full},
	doi = {10.3389/fnins.2018.00941},
	pages = {1--19},
	issue = {December},
	journaltitle = {Frontiers in Neuroscience},
	author = {Knight, James C. and Nowotny, Thomas},
	date = {2018},
}

@inproceedings{comsa2020temporal,
  title={Temporal coding in spiking neural networks with alpha synaptic function},
  author={Comsa, Iulia M and Potempa, Krzysztof and Versari, Luca and Fischbacher, Thomas and Gesmundo, Andrea and Alakuijala, Jyrki},
  booktitle={ICASSP 2020-2020 IEEE International Conference on Acoustics, Speech and Signal Processing (ICASSP)},
  pages={8529--8533},
  year={2020},
  organization={IEEE}
}

@article{huh2018gradient,
  title={Gradient descent for spiking neural networks},
  author={Huh, Dongsung and Sejnowski, Terrence J},
  journal={Advances in neural information processing systems},
  volume={31},
  year={2018}
}

@article{Turner2022,
	title = {{mlGeNN}: accelerating {SNN} inference using {GPU}-enabled neural networks},
	volume = {2},
	copyright = {All rights reserved},
	issn = {2634-4386},
	url = {https://iopscience.iop.org/article/10.1088/2634-4386/ac5ac5},
	doi = {10.1088/2634-4386/ac5ac5},
	number = {2},
	journal = {Neuromorphic Computing and Engineering},
	author = {Turner, James Paul and Knight, James C and Subramanian, Ajay and Nowotny, Thomas},
	month = jun,
	year = {2022},
	note = {Publisher: IOP Publishing},
	pages = {024002},
}

@article{nowotny2014two,
  title={Two challenges of correct validation in pattern recognition},
  author={Nowotny, Thomas},
  journal={Frontiers in Robotics and AI},
  volume={1},
  pages={5},
  year={2014},
  publisher={Frontiers Media SA}
}

@article{Yavuz2016,
	title = {{GeNN}: a code generation framework for accelerated brain simulations},
	volume = {6},
	issn = {2045-2322},
	url = {https://www.nature.com/articles/srep18854},
	doi = {10.1038/srep18854},
	pages = {18854},
	number = {1},
	journaltitle = {Scientific Reports},
	author = {Yavuz, Esin and Turner, James and Nowotny, Thomas},
	date = {2016-05-07},
	pmid = {26740369},
	note = {Publisher: Nature Publishing Group},
}

@article{knight2021pygenn,
  title={PyGeNN: a Python library for GPU-enhanced neural networks},
  author={Knight, James C and Komissarov, Anton and Nowotny, Thomas},
  journal={Frontiers in Neuroinformatics},
  volume={15},
  pages={659005},
  year={2021},
  publisher={Frontiers Media SA}
}

@inproceedings{bena2024event,
  title={Event-based backpropagation on the neuromorphic platform SpiNNaker2},
  author={B{\'e}na, Gabriel and Wunderlich, Timo and Akl, Mahmoud and Vogginger, Bernhard and Mayr, Christian and Gonzalez, Hector Andres},
  booktitle={NeurIPS 2024 Workshop Machine Learning with new Compute Paradigms}
}

@software{genn_5_1_0,
  author       = {James Knight and
                  Thomas Nowotny and
                  James Paul Turner and
                  Esin Yavuz and
                  Fawad Ali and
                  Mengchi Zhang and
                  Anton Komissarov and
                  Ben Evans and
                  Garibaldi Pineda-Garcia and
                  Kanishk Kalra and
                  Alan Diamond and
                  Obaid Ur Rehman and
                  Christoph Ostrau and
                  Alex Cope and
                  Ajay Subramanian and
                  Alex Dewar and
                  Marcel Stimberg and
                  Felix Benjamin Kern and
                  FabianSchubert and
                  Lev E. Givon and
                  Edward Stevinson and
                  Xilin Huang and
                  Anindya Ghosh and
                  Edward Stevinson},
  title        = {genn-team/genn: GeNN 5.1.0},
  month        = nov,
  year         = 2024,
  publisher    = {Zenodo},
  version      = {5.1.0},
  doi          = {10.5281/zenodo.14051978},
  url          = {https://doi.org/10.5281/zenodo.14051978},
}

@software{ml_genn_2_3_0,
  author       = {James Knight and
                  James Paul Turner and
                  Ajay Subramanian and
                  Thomas Nowotny and
                  Isabella Forero and
                  Balázs Mészáros and
                  Adrian D'Alessandro and
                  Jorge Felipe Gaviria and
                  FabianSchubert},
  title        = {genn-team/ml\_genn: ml\_genn\_2\_3\_0},
  month        = dec,
  year         = 2024,
  publisher    = {Zenodo},
  version      = {ml\_genn\_2\_3\_0},
  doi          = {10.5281/zenodo.14258972},
  url          = {https://doi.org/10.5281/zenodo.14258972},
}

@phdthesis{pehle2021adjoint,
  title={Adjoint equations of spiking neural networks},
  author={Pehle, Christian-Gernot},
  year={2021},
  school = {Heidelberg University},
}

@article{shrestha2018slayer,
  title={Slayer: Spike layer error reassignment in time},
  author={Shrestha, Sumit B and Orchard, Garrick},
  journal={Advances in neural information processing systems},
  volume={31},
  year={2018}
}

@article{schuman2017survey,
  title={A survey of neuromorphic computing and neural networks in hardware},
  author={Schuman, Catherine D and Potok, Thomas E and Patton, Robert M and Birdwell, J Douglas and Dean, Mark E and Rose, Garrett S and Plank, James S},
  journal={arXiv preprint arXiv:1705.06963},
  year={2017}
}

@article{james2017historical,
  title={A historical survey of algorithms and hardware architectures for neural-inspired and neuromorphic computing applications},
  author={James, Conrad D and Aimone, James B and Miner, Nadine E and Vineyard, Craig M and Rothganger, Fredrick H and Carlson, Kristofor D and Mulder, Samuel A and Draelos, Timothy J and Faust, Aleksandra and Marinella, Matthew J and others},
  journal={Biologically Inspired Cognitive Architectures},
  volume={19},
  pages={49--64},
  year={2017},
  publisher={Elsevier}
}

@article{thakur2018large,
  title={Large-scale neuromorphic spiking array processors: A quest to mimic the brain},
  author={Thakur, Chetan Singh and Molin, Jamal Lottier and Cauwenberghs, Gert and Indiveri, Giacomo and Kumar, Kundan and Qiao, Ning and Schemmel, Johannes and Wang, Runchun and Chicca, Elisabetta and Olson Hasler, Jennifer and others},
  journal={Frontiers in neuroscience},
  volume={12},
  pages={891},
  year={2018},
  publisher={Frontiers Media SA}
}

@article{isaksson2008cross,
  title={Cross-validation and bootstrapping are unreliable in small sample classification},
  author={Isaksson, Anders and Wallman, Mikael and G{\"o}ransson, Hanna and Gustafsson, Mats G},
  journal={Pattern Recognition Letters},
  volume={29},
  number={14},
  pages={1960--1965},
  year={2008},
  publisher={Elsevier}
}

@article{sun2025towards,
  title={Towards parameter-free attentional spiking neural networks},
  author={Sun, Pengfei and Wu, Jibin and Devos, Paul and Botteldooren, Dick},
  journal={Neural Networks},
  volume={185},
  pages={107154},
  year={2025},
  publisher={Elsevier}
}

@article{muller2022braille,
  title={Braille letter reading: A benchmark for spatio-temporal pattern recognition on neuromorphic hardware},
  author={M{\"u}ller-Cleve, Simon F and Fra, Vittorio and Khacef, Lyes and Peque{\~n}o-Zurro, Alejandro and Klepatsch, Daniel and Forno, Evelina and Ivanovich, Diego G and Rastogi, Shavika and Urgese, Gianvito and Zenke, Friedemann and others},
  journal={Frontiers in Neuroscience},
  volume={16},
  pages={951164},
  year={2022},
  publisher={Frontiers Media SA}
}

@article{walters2025neuromorse,
  title={NeuroMorse: a temporally structured dataset for neuromorphic computing},
  author={Walters, Ben and Bethi, Yeshwanth and Kergan, Taylor and Nguyen, Binh and Amirsoleimani, Amirali and Eshraghian, Jason K and Afshar, Saeed and Rahimi Azghadi, Mostafa},
  journal={Neuromorphic Computing and Engineering},
  year={2025}
}

@article{pedersen2024neuromorphic,
  title={Neuromorphic intermediate representation: A unified instruction set for interoperable brain-inspired computing},
  author={Pedersen, Jens E and Abreu, Steven and Jobst, Matthias and Lenz, Gregor and Fra, Vittorio and Bauer, Felix Christian and Muir, Dylan Richard and Zhou, Peng and Vogginger, Bernhard and Heckel, Kade and others},
  journal={Nature Communications},
  volume={15},
  number={1},
  pages={8122},
  year={2024},
  publisher={Nature Publishing Group UK London}
}

@incollection{bos2023sub,
  title={Sub-mw neuromorphic snn audio processing applications with rockpool and xylo},
  author={Bos, Hannah and Muir, Dylan},
  booktitle={Embedded Artificial Intelligence},
  pages={69--78},
  year={2023},
  publisher={River Publishers}
}

@article{grimaldi2023learning,
  title={Learning heterogeneous delays in a layer of spiking neurons for fast motion detection},
  author={Grimaldi, Antoine and Perrinet, Laurent U},
  journal={Biological Cybernetics},
  volume={117},
  number={4},
  pages={373--387},
  year={2023},
  publisher={Springer}
}

@inproceedings{grappolini2023beyond,
  title={Beyond weights: deep learning in spiking neural networks with pure synaptic-delay training},
  author={Grappolini, Edoardo and Subramoney, Anand},
  booktitle={Proceedings of the 2023 International Conference on Neuromorphic Systems},
  pages={1--4},
  year={2023}
}

@article{jeffress1948place,
  title={A place theory of sound localization.},
  author={Jeffress, Lloyd A},
  journal={Journal of comparative and physiological psychology},
  volume={41},
  number={1},
  pages={35},
  year={1948},
  publisher={American Psychological Association}
}

@inproceedings{grimaldi2022learning,
  title={Learning hetero-synaptic delays for motion detection in a single layer of spiking neurons},
  author={Grimaldi, Antoine and Perrinet, Laurent U},
  booktitle={2022 IEEE International Conference on Image Processing (ICIP)},
  pages={3591--3595},
  year={2022},
  organization={IEEE}
}

@article{wang2019delay,
  title={A delay learning algorithm based on spike train kernels for spiking neurons},
  author={Wang, Xiangwen and Lin, Xianghong and Dang, Xiaochao},
  journal={Frontiers in neuroscience},
  volume={13},
  pages={252},
  year={2019},
  publisher={Frontiers Media SA}
}

@inproceedings{sadovsky2023speech,
  title={Speech command recognition based on convolutional spiking neural networks},
  author={Sadovsky, Erik and Jakubec, Maros and Jarina, Roman},
  booktitle={2023 33rd International Conference Radioelektronika (RADIOELEKTRONIKA)},
  pages={1--5},
  year={2023},
  organization={IEEE}
}

@article{Rozenvasser1967,
        author = {E. N. Rozenvasser},
        title= {General sensitivity equations of discontinuous systems},
journal = {Avtomat. i Telemekh.},
year = {1967},
issue = {3},
pages = {52--56},
comment = {translation: Autom. Remote Control (1967), pp 400--404}
}

@article{galan1999parametric,
  title={Parametric sensitivity functions for hybrid discrete/continuous systems},
  author={Gal{\'a}n, Santos and Feehery, William F and Barton, Paul I},
  journal={Applied Numerical Mathematics},
  volume={31},
  number={1},
  pages={17--47},
  year={1999},
  publisher={Elsevier}
}

@article{Morrison2005,
	title = {Advancing the boundaries of high-connectivity network simulation with distributed computing.},
	volume = {17},
	issn = {0899-7667},
	url = {http://www.ncbi.nlm.nih.gov/pubmed/15969917},
	doi = {10.1162/0899766054026648},
	pages = {1776--801},
	number = {8},
	journaltitle = {Neural computation},
	author = {Morrison, Abigail and Mehring, Carsten and Geisel, Theo and Aertsen, a D and Diesmann, Markus},
	date = {2005-08},
	pmid = {15969917},

}

@article{Brette2007,
	title = {Simulation of networks of spiking neurons: A review of tools and strategies},
	volume = {23},
	issn = {1573-6873},
	url = {http://www.pubmedcentral.nih.gov/articlerender.fcgi?artid=2638500&tool=pmcentrez&rendertype=abstract},
	doi = {10.1007/s10827-007-0038-6},
	pages = {349--398},
	number = {3},
	journaltitle = {Journal of Computational Neuroscience},
	author = {Brette, Romain and Rudolph, Michelle and Carnevale, Ted and Hines, Michael and Beeman, David and Bower, James M and Diesmann, Markus and Morrison, Abigail and Goodman, Philip H and Harris, Frederick C and Zirpe, Milind and Natschläger, Thomas and Pecevski, Dejan and Ermentrout, Bard and Djurfeldt, Mikael and Lansner, Anders and Rochel, Olivier and Vieville, Thierry and Muller, Eilif and Davison, Andrew P and El Boustani, Sami and Destexhe, Alain},
	urldate = {2014-07-11},
	date = {2007-12-12},
	pmid = {17629781},
}

@article{brette_event-driven_nodate,
	title = {Event-driven simulation of integrate-and-fire neurons with exponential synaptic conductances},
	author = {Brette, Romain},
	langid = {english}
}

@article{brette_exact_2007,
	title = {Exact Simulation of Integrate-and-Fire Models with Exponential Currents},
	volume = {19},
	issn = {0899-7667, 1530-888X},
	url = {https://direct.mit.edu/neco/article/19/10/2604-2609/7220},
	doi = {10.1162/neco.2007.19.10.2604},
	pages = {2604--2609},
	number = {10},
	journaltitle = {Neural Computation},
	shortjournal = {Neural Computation},
	author = {Brette, Romain},
	urldate = {2025-06-13},
	date = {2007-10},
	langid = {english}
}

\end{document}